\documentclass[runningheads]{llncs}

\usepackage{eccv}

\usepackage{caption} 
\usepackage{subcaption}
\usepackage{multirow}

\usepackage{eccvabbrv}
\usepackage{floatrow}
\usepackage{graphicx}
\usepackage{amsmath}
\usepackage{amssymb}
\usepackage{ragged2e}
\usepackage{booktabs}
\usepackage{caption} 
\usepackage{multirow}
\usepackage{pifont}
\newcommand{\cmark}{\ding{51}}%
\newcommand{\xmark}{\ding{55}}%

\newcommand{\keypoint}[1]{\noindent\textbf{#1}}
\newcommand{\cut}[1]{}
\usepackage{algpseudocode}

\usepackage[nopar]{lipsum}
\usepackage{arydshln}
\DeclareMathAlphabet\mathbfcal{OMS}{cmsy}{b}{n}
\newcommand{\modelName}[1]{sketch2vec}

\usepackage{makecell}

\usepackage{pgfplots}
\usepackage{tikz}
\pgfplotsset{compat=1.11,
    /pgfplots/ybar legend/.style={
    /pgfplots/legend image code/.code={%
       \draw[##1,/tikz/.cd,yshift=-0.25em]
        (0cm,0cm) rectangle (3pt,0.8em);},
   },
}

\definecolor{deepGreen}{RGB}{0,153,0}
\definecolor{orange}{RGB}{255,125,0}
\definecolor{myblue}{RGB}{0, 129, 207}
\definecolor{mypink}{RGB}{255, 111, 145} 
\definecolor{myblack}{RGB}{75, 68, 83} 
\definecolor{myyellow}{RGB}{255, 150, 113}
\definecolor{sainone}{RGB}{236, 242, 249} %{216, 229, 243}
\definecolor{saintwo}{RGB}{255, 230, 204}

\def\QD#1{\textcolor[HTML]{B85450}{\textbf{#1}}}
\def\TU#1{\textcolor[HTML]{82B366}{\textbf{#1}}}
\def\EM#1{\textcolor[HTML]{6C8EBF}{\textbf{#1}}}
\def\qd#1{\textcolor[HTML]{B85450}{#1}}
\def\tu#1{\textcolor[HTML]{82B366}{#1}}
\def\em#1{\textcolor[HTML]{6C8EBF}{#1}}

\newcommand*\colourcheck[1]{%
  \expandafter\newcommand\csname #1check\endcsname{\textcolor{#1}{\ding{52}}}%
}
\newcommand*\colourcross[1]{%
  \expandafter\newcommand\csname #1cross\endcsname{\textcolor{#1}{\ding{55}}}%
}
\colourcheck{blue}
\colourcheck{green}
\colourcross{red}

\newcommand{\red}[1]{{\color{red}#1}}

\usepackage{wrapfig}

\usepackage[accsupp]{axessibility}  % Improves PDF readability for those with disabilities.

\usepackage[pagebackref,breaklinks,colorlinks,citecolor=eccvblue]{hyperref}
\usepackage{orcidlink}

\usepackage[capitalize]{cleveref}
\crefname{section}{Sec.}{Secs.}
\Crefname{section}{Section}{Sections}
\Crefname{table}{Table}{Tables}
\crefname{table}{Tab.}{Tabs.}

\begin{document}

\title{Do Generalised Classifiers \textit{really work} \\ on Human Drawn Sketches?}

\author{
% \\[-0.6cm]
% \vspace{-0.6cm}
\href{https://hmrishavbandy.github.io/}{Hmrishav Bandyopadhyay}\hspace{.2cm} 
\href{http://www.pinakinathc.me/}{Pinaki Nath Chowdhury}\hspace{.2cm}
\href{https://aneeshan95.github.io/}{Aneeshan Sain} \hspace{.2cm}
\href{https://subhadeepkoley.github.io/}{Subhadeep Koley}\hspace{.2cm}
\href{https://www.surrey.ac.uk/people/tao-xiang}{Tao Xiang}\hspace{.2cm}
\href{https://ayankumarbhunia.github.io/}{Ayan Kumar Bhunia}
\href{https://personalpages.surrey.ac.uk/y.song/}{Yi-Zhe Song}\vspace{-0.3cm} }

\authorrunning{H.~Bandyopadhyay et al.}

\institute{SketchX, CVSSP, University of Surrey \\
\email{\{h.bandyopadhyay, p.chowdhury, a.sain, \\ s.koley, t.xiang, a.bhunia, y.song\}@surrey.ac.uk}
}
\maketitle
\captionsetup{type=figure}
    \hspace{-0.3cm} \begin{minipage}{0.63\linewidth}
        \includegraphics[width=\linewidth]{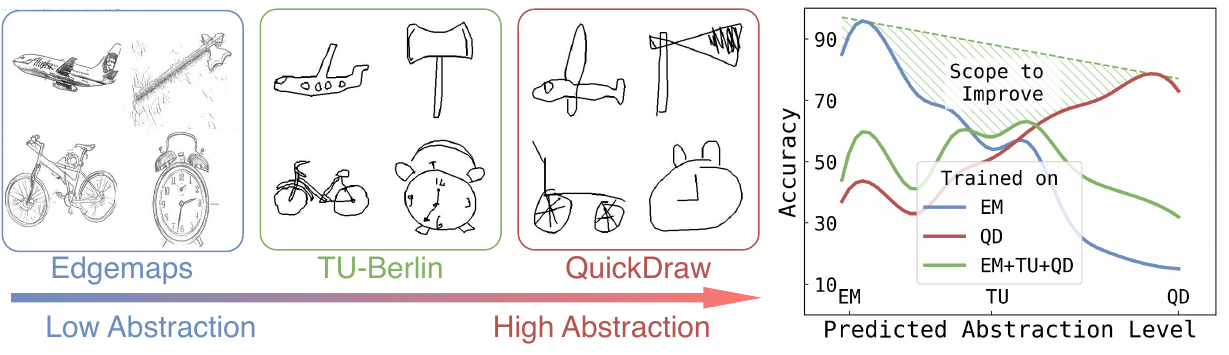}
    \end{minipage}
    \begin{minipage}{0.38\linewidth}
    \fontsize{6}{4}\selectfont 
        Naively training CLIP \(+\) prompt learning on QuickDraw (\qd{QD}), or TU-Berlin (\tu{TU}) sketches, or Edgemaps (\em{EM}) fails to generalise to multiple abstraction levels.\\[0.1cm]
        \begin{tabular*}{\linewidth}{@{\extracolsep{\fill}}cccc}
           \toprule
           
           Training & Evaluation & CLIP & \multicolumn{1}{c}{Ours} \\ \midrule
           \texttt{\qd{QD}} & \texttt{\qd{QD}+\tu{TU}+\em{EM}} & 43.24 & 47.35 ($\uparrow$ 4.1) \\
           \texttt{\tu{TU}} & \texttt{\qd{QD}+\tu{TU}+\em{EM}} & 43.71 & 51.03 ($\uparrow$ 7.3) \\
           \texttt{\em{EM}} & \texttt{\qd{QD}+\tu{TU}+\em{EM}} & 42.91 & 43.76 ($\uparrow$ 0.8) \\
           \texttt{\qd{QD}+\tu{TU}+\em{EM}} & \texttt{\qd{QD}+\tu{TU}+\em{EM}} & 45.59 & 62.96 ($\uparrow$ \textbf{17.4}) \\ \bottomrule
        \end{tabular*} 
    \end{minipage}
  \vspace{-0.2cm}
    \captionof{figure}{Unlike photos, sketch classification poses additional challenges such as \textit{abstraction} -- humans draw differently based on their subjective interpretations, sketching ability, and drawing time. Existing datasets such as TU-Berlin \cite{berlin} and QuickDraw \cite{quickdraw} only capture the time axis and collect human sketches drawn under $280$ and $20$ seconds, respectively. Following \cite{berlin, quickdraw, hertzmann2020line, vinker2022clipasso} we consider Edgemaps (\EM{EM}) as low abstract, TU-Berlin (\TU{TU}) sketches as medium abstract, and QuickDraw (\QD{QD}) ones as highly abstract (\textit{left}). Naively training CLIP via prompt learning, \cite{zhou2022learning} on sketches (\textit{right}) from one abstraction level (\QD{QD}, \TU{TU}, or \EM{EM}) individually do not generalise across varying abstractions (\QD{QD} + \TU{TU} + \EM{EM}). Jointly training on multiple abstractions (\QD{QD} + \TU{TU} + \EM{EM}) is also sub-optimal ($45.6$ on CoOp \cite{zhou2022learning} vs $62.9$ on Ours). (\textit{middle}) Importantly, our proposed method predicts a classification score and an abstraction level for input sketches. Plotting classification accuracy vs predicted abstraction level, reveals a \textit{scope for improvement} (shaded region) showing -- despite our significant improvement in classification ($\uparrow$ 17.4\%) over naive CLIP + prompt learning, generalisation across varying sketch abstractions is still an open problem. We hope this will motivate future works to democratise existing methods \cite{clip, zhou2022conditional} for human drawn sketches.}
    \label{fig:teaser}
    
\begin{abstract}
    \vspace{-0.5cm}
    
    This paper, for the first time, marries large foundation models with human sketch understanding. We demonstrate what this brings -- a paradigm shift in terms of generalised sketch representation learning (e.g., classification). This generalisation happens on two fronts: (i) generalisation across unknown categories (i.e., open-set), and (ii) generalisation traversing abstraction levels (i.e., good and bad sketches), both being timely challenges that remain unsolved in the sketch literature. Our design is intuitive and centred around transferring the already stellar generalisation ability of CLIP to benefit generalised learning for sketches. We first ``condition'' the vanilla CLIP model by learning sketch-specific prompts using a novel auxiliary head of raster to vector sketch conversion. This importantly makes CLIP ``sketch-aware''. We then make CLIP acute to the inherently different sketch abstraction levels. This is achieved by learning a codebook of abstraction-specific prompt biases, a weighted combination of which facilitates the representation of sketches across abstraction levels -- low abstract edge-maps, medium abstract sketches in TU-Berlin, and highly abstract doodles in QuickDraw. Our framework surpasses popular sketch representation learning algorithms in both zero-shot and few-shot setups and in novel settings across different abstraction boundaries.
    
\end{abstract}

\vspace{-0.6cm}
\section{Introduction}
\vspace{-0.3cm}
\noindent The vision community is witnessing a paradigm shift in the face of large foundation models \cite{clip, LDM}. Instead of learning visual semantics from scratch, the rich semantics inherent to foundation models are explored to enrich visual learning, as in \cite{baldrati2022effective, fang2021clip2video, lei2021less} for retrieval, and \cite{text2light2022, mirowski2022clip, wang2022clip} for generation. The most salient advantage that foundation models bring is their generalisation ability \cite{zhou2022learning, CLIP-Adapter, zhou2022conditional, khattak2022maple}, which made a significant impact on zero-shot and few-shot learning.

In this paper, we marry human sketches with foundation models {to naturally tackle two of most significant bottlenecks in the sketch community, with little effort, piggybacking on generalisability of foundation models.} First is the data scarcity problem of human drawn sketches -- the largest sketch dataset (QuickDraw \cite{quickdraw}) contains $350$ categories compared with the easily $>1000$ categories for photos \cite{deng2009imagenet}, which makes generalised learning for sketch even more pronounced a problem. The second is, although everyone \textit{can} sketch, most people sketch \textit{differently} as a result of varying drawing skills and diverse subjective interpretations -- see \cref{fig:teaser} for distinctly different sketches of a ``bike''. These sketch-specific challenges call for a single model generalising along two axes -- (i) across \textit{unseen categories} for the first data scarcity challenge, and (ii) across \textit{abstraction levels} for the second challenge of sketches exhibiting varying abstraction levels. 

Solving these challenges, we show that it all comes down to making CLIP \cite{clip} sketch-specific. For the former (data scarcity problem), we learn a set of continuous vectors (visual prompts) injected into CLIP's vision transformer encoder. This enables CLIP (pre-trained on $\sim$$400$M image-text pairs) to adapt to sketches while preserving its generalisation ability -- resulting in a generalised sketch recogniser that works across unknown categories. More specifically, we first design two sets of \textit{visual prompts} -- shallow prompts injected into the initial layer of the CLIP transformer and deep prompts injected up to a depth of $9$ layers. Keeping rest of CLIP frozen, we train these prompts on the extracted sketch and \texttt{[category]} embeddings (as class labels) from CLIP text encoder using cross-entropy loss. Although shallow+deep prompts encourage CLIP to become ``sketch-aware'', they do not model any sketch-specific traits during training. Hence, we additionally use an auxiliary task of raster to vector sketch conversion that exploits the dual representation of sketch \cite{sketch2vec} to reinforce that awareness.

The latter challenge of dealing with sketch abstraction\footnote{For rest of the paper, we denote `abstraction' as $\mathbb{A}$.} is less obvious and extends the status quo of what is possible with foundation models. {While there is no consensus on \textit{what constitutes an abstract sketch} \cite{das2021sketchode}, prior works follow: (i) number of strokes (more stroke $\rightarrow$ less abstract) \cite{vinker2022clipasso}, and (ii) sketch drawing time (more time $\rightarrow$ less abstract) \cite{quickdraw, berlin}. Since precisely annotating the abstraction score for each sketch is ill-posed, we learn sketch abstraction semi-supervised. Particularly, we assign semi-accurate, coarse abstraction levels as: (i) doodles from QuickDraw dataset \cite{quickdraw} drawn under $20$ seconds as \textit{high abstraction}, (ii) freehand sketches from TU-Berlin \cite{berlin} drawn under $280$ seconds as \textit{medium abstraction}, and (iii) Edgemaps from \cite{zhang2016sketchnet} as \textit{low abstraction}. To model the continuous abstraction spectrum (low $\rightarrow$ high), we employ codebook-learning \cite{VQ-VAE2017}. Our abstraction codebook comprises of three ``codes'' (a learned feature embedding), where the continuous abstraction is modelled by mixing (a weighted average) the discrete codes. To make CLIP ``abstraction-aware'', the abstraction vector (from mixing codes) is injected as an additional prompt to the generalised sketch classifier. To our best knowledge, ours is the first work showing the potential of combining codebook learning \cite{VQ-VAE2017} and prompt learning \cite{zhu2022prompt} for generalised sketch representation learning.}

In summary, our contributions are: (i) we, for the first time marry human sketches with foundation models to tackle two of the most significant bottlenecks facing the sketch community -- data scarcity and abstraction levels. (ii) For data scarcity, we achieve generalisation across unseen categories by adapting CLIP for sketch classification via prompt learning. (iii) To further make CLIP ``sketch-aware'', we exploit sketch-specific traits like raster-to-vector sketch conversion as an auxiliary loss. (iv) For abstraction, we achieve generalisation across varying abstraction levels  using a codebook-driven approach, where a mixup of learned codebook vectors acts as prompts that interface with CLIP, with CLIP, to make our model robust to recognising sketches from multiple abstraction levels, including those not seen during training.
\vspace{-0.4cm}
\section{Related Works}
\vspace{-0.2cm}
\keypoint{Sketch for Visual Understanding:} Sketches can depict visual concepts \cite{hertzmann2020line} using intuitive free-hand drawings, overcoming linguistic barriers often faced in text representations. Fine-grained in nature, a sketch is an attractive query medium for tasks like sketch-based image \cite{dutta2019semantically,sain2021stylemeup,collomosse2019livesketch, shen2018zero} and 3D shape retrieval \cite{xu2022domain}. Creative sketches \cite{ge2021creative} encouraged sketch-based synthesis and editing of images \cite{zeng2022sketchedit, liu2021deflocnet, yu2019free}, natural objects or scenes \cite{chen2018sketchygan,gao2020sketchycoco, wang2021sketch}, faces\cite{chen2020deepfacedrawing}, and animation frames \cite{yi2022animating}. Despite being expressed as monochrome lines on a 2D plane, sketches convey complex 3D structures and find use in 3D shape modelling \cite{guillard2021sketch2mesh,zhang2021sketch2model,yan2020interactive, berardi2023clip}. As an interactive medium \cut{in a limited number of samples}, a sketch is an important modality of input in vision tasks like sketch-based object detection \cite{tripathi2020sketch}, image-inpainting \cite{yu2019free}, representation learning \cite{wang2021sketchembednet}, incremental learning \cite{bhunia2022doodle}, image-segmentation \cite{hu2020sketch}, etc. Beyond its discriminative \cite{sain2021stylemeup} or representative \cite{wang2021sketchembednet} potential, sketch has also been employed for pictionary style gaming \cite{bhunia2020pixelor}. The success of sketch-based visual understanding leads us to propose a framework for recognising \emph{any} (open-set \cite{open-world-recognition-2015}) free-hand drawing on unseen categories.

\vspace{0.1cm}
\keypoint{Sketch Classification:} Early approaches to sketch understanding \cite{sezgin2001interface} extract hand-crafted features like shape primitives \cite{paleosketch}, bag-of-words \cite{berlin}, or Fischer Vectors \cite{schneider2014sketch} from raster (static pixel-map) sketches. Better representations were formulated with deep learning algorithms, dramatically improving sketch classification on complete \cite{ravi2015recognition, deepsketch2015} and partial sketches \cite{deepsketch2016}, surpassing even humans \cite{yu2015sketch} in recognition. Alternate architectural designs encode the temporal order of vector strokes (pen coordinates \& motions) \cite{quickdraw} using RNNs \cite{quickdraw} or Transformers \cite{ribeiro2020sketchformer}. Fusion of both raster representations for encoding structural information and vector representations for temporal cues has been employed with CNN-RNN feature fusion \cite{hanhui_multistage} for million-scale recognition with hashing \cite{xu2018sketchmate}, or with Graph Neural Networks \cite{lan2020S3Net, xu2022multigraph, sketchhealer, sketchlattice}. Various sketch-specific self-supervised methods have recently emerged that employ BERT-like \cite{devlin2018bert} pre-training \cite{lin2020sketch}, jigsaw solving \cite{sketch-jigsaw}, or cross-modal translation between raster-vector dual representation of sketches \cite{sketch2vec}. While recognition in existing works is limited to seen classes, we, for the first time, introduce a zero-shot recognition pipeline that can recognise unseen classes under varying sketch abstraction levels.

\vspace{0.1cm}
\keypoint{CLIP in Vision Tasks:} Contrastive Language-Image Pre-training (CLIP) \cite{clip} pairs rich semantics from text-descriptions with large-scale ($\sim$$400$M) image datasets, exploiting underlying relationships in cross-modal data (image+text) by representing them in the same embedding space. As such, CLIP features are highly generalisable in downstream tasks using very-few (few-shot) or no labels (zero-shot) \cite{zhou2022learning} as opposed to traditional features (trained on discrete labels). This adaptation to downstream tasks like few-shot recognition \cite{CLIP-Adapter}, retrieval \cite{baldrati2022effective, sain2023clip}, object detection \cite{gu2022openvocabulary}, semantic segmentation \cite{DenseCLIP}, image generation \cite{text2light2022}, etc., is done primarily through prompting, first introduced for NLP \cite{brown2020language}. Prompting, in general, involves construction of a task-specific template (e.g., \texttt{`[MASK] is the capital of the France'}), which is then filled with word labels (e.g. \texttt{`London/Paris'}). Prompts give context to the word labels, forming an appropriate text description. Engineering prompts by hand requires domain expertise, hence recent works (prompt tuning) model them as learnable continuous vectors \cite{zhou2022conditional, zhou2022learning} to be optimised directly during fine-tuning. Beyond language prompts, visual prompts in the form of continuous vectors have also been explored in visual feature extractors like ViT \cite{jia2022visual}. In contrast to learning general \cite{zhou2022learning,jia2022visual,bahng2022visual} or instance-specific prompts \cite{zhou2022conditional}, we learn continuous abstraction prompts modelled on abstraction level of input sketch. This enables the recognition of a wide range of sketches, from professional edge-map like drawings to free-hand abstract doodles \cite{quickdraw}. 

\vspace{0.1cm}
\keypoint{Abstraction in Sketches:} Sketch abstraction ($\mathbb{A}$) was first defined as a \emph{factor of time} \cite{berger2013style} with an observation that users tend to draw only salient regions in a constrained time setting. This idea was later modelled as a \emph{trade-off} between \emph{compactness and recognisability} \cite{muhammad2018learning, muhammad2019goal} for the `generation' of abstract strokes by removing strokes based on their salience. Parametric representations model abstraction through B\'ezier curves, \cite{das2020beziersketch} and differential equations \cite{das2021sketchode}, controlling stroke ``complexity'' with B\'ezier control points and sinusoidal frequencies respectively. Alternate representations of abstraction include modelling sketch as a combination of appearance and structure \cite{yang2021sketchaa} in a coarse-to-fine manner through hierarchical feature encoder learning \cite{sain2020cross}, or via a composition of predefined drawing primitives \cite{alaniz2022abstracting} to identify the most distinctive parts of the sketch, and ground them into interpretable features. In contrast, we represent abstraction on a continuous spectrum, varying from Edgemaps (equivalent to professional sketches) to human-drawn doodles (highly abstract) by learning abstraction-specific codebook vectors, which we interpolate in a zero/few-shot sketch recognition setup. 

\vspace{-0.4cm}
\section{Background}
\label{sec: background}
\vspace{-0.3cm}

\keypoint{CLIP:} The generalisability of CLIP makes it a popular choice \cite{sain2023clip} for open-set vision-language tasks. Specifically, CLIP uses $2$ independent encoders: \textit{(i)} a ResNet \cite{he2016deep} or Vision Transformer \cite{dosovitskiy2021image} image encoder, and (ii) a transformer-based \cite{vaswani2017attention} text encoder. The Vision Transformer image encoder $F$ processes input images as $r$ fixed-sized patches $\mathrm{I} = \{p_{1}, \dots, p_{r}\};$ $\ p_{j} \in \mathbb{R}^{3 \times h \times w}$ that are embedded along with an extra learnable class token \cite{devlin2018bert} $c^{p}$. These are then passed through transformer layers \cite{vaswani2017attention} with multi-head attention to obtain the visual features $f_p = F(I, c^{p}) \in \mathbb{R}^{d}$. The text input (say, $n$ words) is pre-processed to word-embeddings $W_{0} = \{\mathbf{w}_{0}^{j}\}_{j=1}^{n}$, from which the text transformer $T$ extracts textual features as $t_0 = \mathrm{T}(W_{0}) \in \mathbb{R}^{d}$. Since CLIP maps cross-modal (image+text) features on the same embedding space, features of text-photo pairs have maximal similarity $\texttt{sim}(\cdot)$ compared to features from unpaired (mismatched) samples after contrastive training. For zero-shot classification, textual prompts like `\texttt{a photo of a [category]}' (from a list of $K$ categories) are used to obtain category-specific textual features $\{t_{j}\}_{j=1}^{K}$. The probability of input photo $I$ (with photo-feature $f_p$) belonging to $y^{\text{th}}$ category can be calculated as
\vspace{-0.2cm}
\begin{equation}\label{eq:clip-loss}
    \mathbb{P}(y | \mathrm{I}) = \frac{\exp\;( \texttt{sim}\;(f_{p}, t_{y}) / \tau )}{ \sum_{j=1}^{K} \exp\;( \texttt{sim}\;(f_{p}, t_{j}) / \tau) }
\vspace{-0.1cm}
\end{equation}

\keypoint{Prompt Learning:} Prompt learning uses foundation models, like BERT \cite{devlin2018bert} and CLIP \cite{clip}, as knowledge bases from which useful information can be extracted for downstream tasks \cite{petroni2019prompt}. Apart from using handcrafted prompts like `\texttt{a photo of a}', recent methods like CoOp \cite{zhou2022learning} and CoCoOp \cite{zhou2022conditional} learns $n$ continuous context vectors, $\{v_{1}, \dots, v_{n}\}$, each having $v_{j} \in \mathbb{R}^{d_{t}}$ dimension word embeddings. With base CLIP frozen, the continuous context vectors $v_{j}$ are learned by backpropagating gradients through the text $\mathrm{T}(\cdot)$ encoder. Using word embedding for the $k$-th `\texttt{[category]}', denoted as ${w_{0}^{k}} \in \mathbb{R}^{d_{t}}$, the prompt is constructed as $[v_{1}, \dots, v_{n}, {w_{0}^{k}}]$, and passed to the transformer to obtain text feature ${f}_{t} = \mathrm{T}([v_{1}, \dots, v_{n}, {w_{0}^{k}}])$. Recent works learn continuous context vectors (i.e., prompts) for text \cite{zhou2022learning} and image \cite{bahng2022visual} multimodal prompts \cite{khattak2022maple}. Variations of prompt learning also include shallow vs. deep \cite{xing2022class} prompts, depending on the layers where the context vectors are injected. We use {deep} prompt learning to adapt CLIP to design a generalised sketch classifier.

\vspace{-0.5cm}
\section{Proposed Methodology}
\vspace{-0.3cm}

In this paper, we build a generalised sketch classifier that works in an unseen setup. This ``unseen'' problem in sketch representation learning has two axes: (a) generalisation across unseen categories --  train on `cats' or `dogs' (not on `zebras') but evaluate on `zebras'; and (b) generalisation across abstractions -- a sketch can be drawn in $20$ seconds (i.e., highly abstract doodle) or $280$ seconds (TU-Berlin sketches \cite{berlin}). For generalisation across categories, we use the open-vocabulary potential of CLIP \cite{clip}, which has excellent generalisation across several downstream tasks. Particularly, we show how off-the-shelf CLIP is sub-optimal, and a simple yet significant sketch-specific adaptation 
with prompt learning \cite{zhu2022prompt} and raster-to-vector self-reconstruction objective \cite{sketch2vec} can help generalisation to unseen categories. Generalising across abstraction ($\mathbb{A}$) levels is challenging as $\mathbb{A}$ is hard-to-label and a \textit{subjective metric} (e.g., it is hard quantifying ``how abstract is that sketch'' on a scale $0$ to $1.0$). Hence, we propose a weakly-supervised codebook-learning paradigm \cite{VQ-VAE2017} to learn generalisation across sketch abstractions.

\vspace{-0.4cm}
\subsection{Generalisation Across Unseen Categories}
\label{sec: generalise-across-categories}
\vspace{-0.2cm}

\keypoint{Baseline Sketch Classifier:} {Sketch classification aims at predicting the category a given query-sketch belongs to.} An input raster sketch $\mathrm{I}_{S} \in \mathbb{R}^{3 \times H \times W}$ is encoded using a backbone feature extractor $f_{s} = \mathrm{F}(\mathrm{I}) \in \mathbb{R}^{d}$ like ResNet-101 \cite{he2016deep} followed by {mapping it to a $K$-dimensional vector} $\mathrm{F}_{c}: \mathbb{R}^{d} \rightarrow \mathbb{R}^{K}$ that classifies $\mathrm{I}_{S}$ into predefined $K$ categories $f_{c} = \mathrm{F}_{c}(f_{s}) \in \mathbb{R}^{K}$. Both backbone $\mathrm{F}(\cdot)$ and classifier $\mathrm{F}_{c}(\cdot)$ are learned given ground-truth class $\hat{f}_{c,k}$ as,
\vspace{-0.2cm}
\begin{equation}
    \mathcal{L}_\text{CE} = - \hat{f}_{c, k} \log \frac{ \exp(f_{c, k}) }{ \sum_{j=1}^{K} \exp (f_{c, j}) }
    \label{eq:eq2}
\end{equation}
\keypoint{Prompt Learning to Adapt CLIP for Sketches:} 
We use CLIP with ViT visual encoder~\cite{clip} which {extends the fixed set classifier $\mathrm{F}_{c}$ in \cref{eq:eq2} into an open-set setup}. We now learn $J$ sketch prompts $\mathbf{v^{s}} = \{v^{s}_{0}, \dots, v^{s}_{J-1}\}$, where $v^{s}_{j} \in \mathbb{R}^{5 \times d_{p}}$. First, we divide the raster sketch into $r$ fixed-sized patches $\mathrm{I}_{S} = \{s_{1}, \dots, s_{r}\}$ where each patch $s_{i} \in \mathbb{R}^{3 \times h \times w}$ is embedded as $E_{0} = \{e_{0}^{j}\}_{j=1}^{r}$. Next, the learnable prompts are injected into each transformer block of CLIP vision transformer $\mathrm{F}(\cdot)$ up to a specific depth $J$, as
\vspace{-0.2cm}
\begin{equation}
    \begin{split}
        \hspace{-0.4cm} [c^{p}_{j}, E_{j}, \textbf{\underline{\textcolor{white}{a}}}\;] &= \mathrm{F}_{j} ([c^{p}_{j-1}, E_{j-1}, v^{s}_{j-1}]) 
        \; |_{j=1}^J \\
        \hspace{-0.4cm} [c^{p}_{i}, E_{i}, v^{s}_{i}] &= \mathrm{F}_{i} ([c^{p}_{i-1}, E_{i-1}, v^{s}_{i-1}])
        \; \, \, |_{i=J+1}^N \\
        f_{s} &= \texttt{ImageProj} (c^{p}_{N}) \\[-0.05cm]
    \end{split}
\end{equation}
\noindent where, $c^{p}_{0}$ is a pre-trained \texttt{[CLS]} token (see \cref{sec: background}). To classify the visual feature $f_{s} \in \mathbb{R}^{d}$, we use handcrafted prompts like `\texttt{a photo of a [category]}' that is encoded using CLIP text encoder $\mathrm{T}(\cdot)$ into $f_{t}$ as in \cref{eq:clip-loss}. However, (i) our input is `\texttt{sketch}' not `\texttt{photo}', and (ii) handcrafted prompts are sub-optimal compared to learnable prompts \cite{zhou2022learning}, $\mathbf{v^{t}} = \{v^{t}_{0}, \dots, v^{t}_{J-1}\}$; $v^{t}_{j} \in \mathbb{R}^{5 \times d_{t}}$. Hence, we inject prompts $\mathbf{v^{t}}$ in $\mathrm{T}(\cdot)$ up to depth $J$ as,
\vspace{-0.2cm}
\begin{equation}
    \begin{split}
        [\;\textbf{\underline{\textcolor{white}{a}}}\;, {w}_{j}^{{k}}] &= \mathrm{T}_{j}([v^{t}_{j-1}, {w}_{j-1}^{{k}}])
        \; |_{j=1}^J \\
        [v^{t}_{i}, {w}_{i}^{{k}}] &= \mathrm{T}_{i}([v^{t}_{i-1}, {w}_{i-1}^{{k}}])
        \; \, \, |_{i=J+1}^N \\
        f_{t} &= \texttt{TextProj} ({w}_{N}^{{k}}) \\[-0.1cm]
    \end{split}
\end{equation}
where, $w^{k}_{0}$ is the word embedding of `\texttt{[category]}'. Naively using learnable prompts $\mathbf{v^{t}}$ overfits to training/seen categories, lacking generalisation to unseen categories \cite{zhou2022conditional}.
Hence, we use a lightweight Meta-Net $\pi = \mathrm{H}(f_{s})$ to predict an instance-specific context, $\pi \in \mathbb{R}^{5 \times d_{t}}$ that shifts $\mathbf{v^{t}}$ as $\mathbf{v^{t}}(f_{s}) = \{v^{t}_{0} + \pi, \dots, v^{t}_{J-1}+\pi \}$. Intuitively, Meta-Net (a two-layer Linear-ReLU-Linear) reduces overfitting of $\mathbf{v^{t}}$ to training/seen categories, generalising better for unseen classes using sketch-conditional prompts $\mathbf{v^{t}}(f_{s})$. Finally,
\vspace{-0.1cm}

\begin{equation}\label{eq:clip-loss-meta-net}
    \mathcal{L}_\text{CE} = -\log \frac{\exp \; ( \texttt{sim} \; (f_{s}, \mathrm{T}([\mathbf{v^{t}}(f_{s}), w^{i}_{0}]) \; ) / \tau )}{ \sum_{j=1}^{K} \exp \; ( \texttt{sim} \; (f_{s}, \mathrm{T}([\mathbf{v^{t}}(f_{s}), w^{j}_{0}]) \; ) / \tau ) }
\vspace{0.1cm}
\end{equation}

\keypoint{Auxiliary Loss using Sketch Specific Traits:} Sketches are uniquely characterised by its existence in dual modalities -- rasterised images $\mathrm{I}_{S} \in \mathbb{R}^{3 \times H \times W}$ and vector coordinate sequences $\mathrm{I}_{V} \in \mathbb{R}^{N \times 5}$. Translating $\mathrm{I}_{S} \rightarrow \mathrm{I}_{V}$ enforces image encoder $\mathrm{F}(\cdot)$, particularly its learnable prompts $\mathbf{v^{s}}$, to learn sparse stroke information. Accordingly, we use a linear embedding layer \cite{sketch2vec} to obtain the initial hidden state of a Gated Recurrent Unit (GRU) decoder as $h_{0} = W_{h} f_{s} + b_{h}$. Its hidden state $h_{t}$ is updated as: $h_{t} = \texttt{GRU}(h_{t-1}; [f_{s}, P_{t-1}])$, where $P_{t-1}$ is the last predicted point. Next, an embedding layer is used to predict a five-element vector at each time step as $P_{t} = W_{p}h_{t} + b_{p}$ where $P_{t} = (x_{t}, y_{t}, q_{t}^{1}, q_{t}^{2}, q_{t}^{3}) \in \mathbb{R}^{2+3}$ whose first two logits represent absolute coordinate $(x,y)$ and the later three for pen's state position $(q^{1}, q^{2}, q^{3})$. We use mean-square error and categorical cross-entropy loss to train raster $\to$ vector on ground-truth absolute coordinates $(\hat{x}_{t}, \hat{y}_{t})$ and pen state ($\hat{q}_{1}, \hat{q}_{2}, \hat{q}_{3}$) as,
\vspace{-0.3cm}
\begin{equation}\label{eq: raster-vector}
\begin{split}
    \mathcal{L}_{\text{S} \to \text{V}} &= \frac{1}{T} \sum_{t=1}^{N} ||\hat{x}_{t} - x_{t}||_{2} + ||\hat{y}_{t} - y_{t}||_{2} \\[-2pt]
    & - \frac{1}{N} \sum_{t=1}^{N} \sum_{i=3}^{3} \hat{q}_{t}^{i} \log \Big( \frac{\exp(q_{t}^{i})}{\sum_{j=1}^{3} \exp (q_{t}^{j})} \Big) \\[-0.4cm]
\end{split}
\vspace{-0.7cm}
\end{equation}

\vspace{-0.3cm}
\subsection{Generalisation Across Sketch Abstractions}
\label{sec: abstraction}
\vspace{-0.1cm}

\keypoint{Pilot Study:} Here we elaborate \cref{fig:teaser} (\textit{right}) that examines generalisation of CLIP when abstractions vary from \EM{EM} $\to$ \TU{TU} $\to$ \QD{QD}. We randomly select $40$ classes common across \QD{QD}, \TU{TU} and \EM{EM} -- $20$ seen classes to adapt CLIP via prompt learning (CoOp \cite{zhou2022learning}), and $20$ unseen classes for zero-shot evaluation. We observe: (i) training and evaluating on the same abstraction (\QD{QD}, or \TU{TU}, or \EM{EM}) performs {$\sim$$5.09\%$} better, than training on one and jointly evaluating on \QD{QD} + \TU{TU} + \EM{EM}. This drop signifies that a naive CLIP + prompt learning fails to generalise across abstractions. (ii) Jointly training on sketches from \QD{QD} + \TU{TU} + \EM{EM}, only marginally improves accuracy by {$2.30\%$}. This highlights an \textit{even deeper} problem -- simply scaling the training data will \textit{likely}\footnote{Exploring scaling laws \cite{scaling-laws} for human sketch abstraction is an interesting and non-trivial problem for future work.} not solve the \textit{varying} sketch abstraction problem.

\vspace{0.1cm}
\keypoint{Overview:} Although we can easily sketch at varying abstraction levels, collecting precise abstraction annotation is difficult. \cref{fig:teaser} (\textit{left}), shows that \textit{in general}, \QD{QD} doodles are more abstract than \TU{TU} sketches, however precisely annotating abstraction score from $0 \to 1$ for every sample (e.g., ``bike'' in \QD{QD}) is an ill-posed problem. Prior works developed \textit{proxy metrics} for sketch abstraction, like the number of strokes (fewer strokes $\Rightarrow$ higher abstraction) \cite{vinker2022clipasso}, drawing skills (amateur vs. professional) \cite{gryaditskaya2019opensketch}, and time to sketch (lesser time $\Rightarrow$ higher abstraction) \cite{sangkloy2016sketchy}. Instead, we design our method based on the general consensus \cite{berlin, hertzmann2020line, quickdraw, sangkloy2016sketchy} that -- (i) \EM{EM}s are (usually) less abstract than \TU{TU}, and (ii) \QD{QD} doodles are (usually) more abstract than \TU{TU}. Particularly, while \QD{QD} + \TU{TU} + \EM{EM} do not cover all possible sketch abstractions, \EM{EM} and \QD{QD} are a good approximation of the \textit{lower} and \textit{upper} bounds of sketch abstractions.

\setlength{\intextsep}{1.5pt}
\begin{wrapfigure}{r}{0.5\textwidth}
\includegraphics[width=\linewidth]{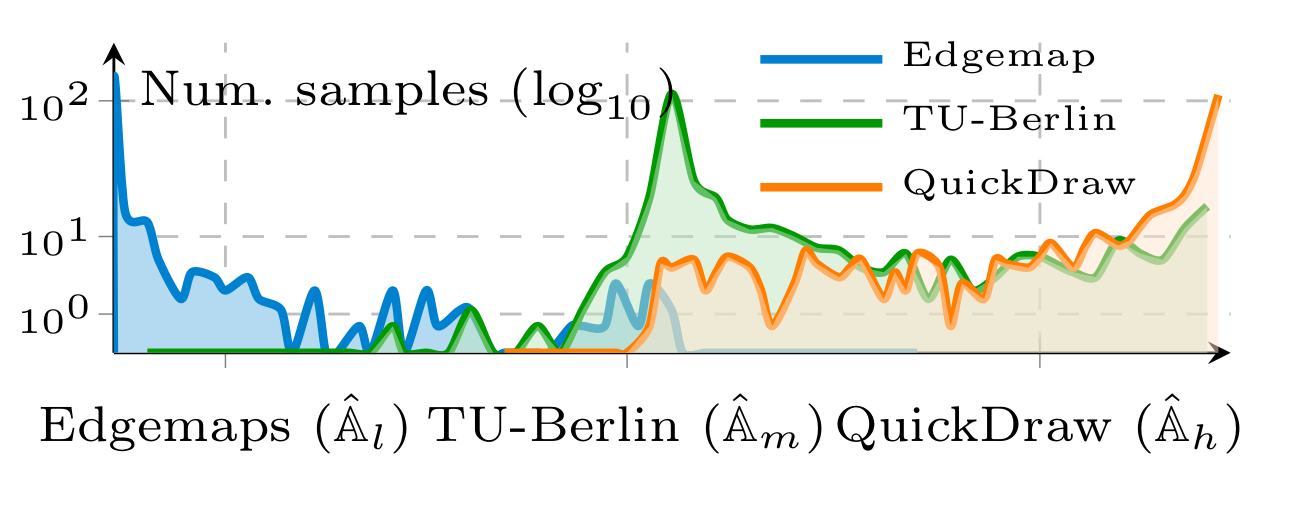}
\vspace{-0.8cm}

\caption{Plotting number of sketches vs class membership of $600$ sketch instances, defined by class-labels ($\hat{\mathbb{A}}_{l}, \hat{\mathbb{A}}_{m}, \hat{\mathbb{A}}_{h}$) and softmax normalised distributions ($\mathbb{A}_{l}, \mathbb{A}_{m}, \mathbb{A}_{h}$). Sketches are taken from $20$ unseen categories common across \QD{QD}, \TU{TU}, and \EM{EM} with $10$ sketches per category. Despite expected peaks, a significant number of sketches lie in the continuous spectrum between $\hat{\mathbb{A}}_{l} \to \hat{\mathbb{A}}_{m}$ and $\hat{\mathbb{A}}_{m} \to \hat{\mathbb{A}}_{h}$ (overlaps).}
\label{fig: expected-abstraction}
\vspace{-0.15cm}
\end{wrapfigure}

\vspace{0.1cm}
\keypoint{Abstraction ($\mathbb{A}$) Learning without Annotations:} Although drawing a sketch at varying abstractions is easy, annotating its precise abstraction level is hard, inferring the need of a weakly-supervised approach for abstraction modelling. Given that \EM{EM} $\to$  \QD{QD} roughly provides a low$\to$high abstraction \cite{berlin, hertzmann2020line, quickdraw, sangkloy2016sketchy}, we define a classification problem: represent \QD{QD} (high $\mathbb{A}$) by ground-truth class label $\hat{\mathbb{A}}_{h}$, \TU{TU} (medium $\mathbb{A}$) by $\hat{\mathbb{A}}_{m}$, and \EM{EM} (low $\mathbb{A}$) by $\hat{\mathbb{A}}_{l}$ . For each abstraction level, we learn a codebook vector $\{ \theta_{l}, \theta_{m}, \theta_{h} \}$ where $\theta_{l,m,h} \in \mathbb{R}^{5 \times d_{t}}$. Given an input sketch $\mathrm{I}_{S}$, our backbone sketch encoder extracts $f_{s} = \mathrm{F}(\mathrm{I}_{S}, \mathbf{v^{s}})$, which is then fed into a \emph{codebook classifier} $\mathcal{C}_{\theta}: \mathbb{R}^{d} \rightarrow \mathbb{R}^{3}$ to get a \texttt{softmax} normalised probability distribution over the three abstraction class labels as, $[\mathbb{A}_{l}, \mathbb{A}_{m}, \mathbb{A}_{h}] = \mathcal{C}_{\theta}(f_{s})$. We train $\mathcal{C}_\theta(\cdot)$ via a categorical cross-entropy loss as,

\vspace{-0.3cm}
\begin{equation}\label{eq: codebook-classify}
    \mathcal{L}_\text{CB} = - (\hat{\mathbb{A}}_{l} \log \mathbb{A}_{l} + \hat{\mathbb{A}}_{m} \log \mathbb{A}_{m} + \hat{\mathbb{A}}_{h} \log \mathbb{A}_{h})
\vspace{-0.1cm}
\end{equation}

\noindent The predicted scores are used to combine (weighted summation) learned codebooks as $\eta = \mathbb{A}_{l} \theta_{l} + \mathbb{A}_{m} \theta_{m} + \mathbb{A}_{h} \theta_{h}$ which acts as an abstraction prompt $\eta \in \mathbb{R}^{5 \times d_{t}}$ and shifts the sketch-conditional prompt $\mathbf{v^{t}}(f_{s})$ (\cref{sec: generalise-across-categories}) like a bias as, $\mathbf{v^{t}}(f_{s}^{*}) = \{(v_{0}^{1} + \pi + \eta), \dots, (v_{J-1}^{t} + \pi + \eta) \}$. Next, we use \cref{eq:clip-loss-meta-net} to classify.

\vspace{0.1cm}
\keypoint{Augmentations using Abstraction-Mixup:} \cref{eq: codebook-classify} helps us model abstractions in human drawn sketch by learning a codebook vector for $3$-levels $[\theta_{l}, \theta_{m}, \theta_{h}]$ and predicting their \texttt{softmax} normalised probabilities $[\mathbb{A}_{l}, \mathbb{A}_{m}, \mathbb{A}_{h}]$. Unseen sketches however do not adhere to these predefined levels and are often on a continuous \textit{spectrum} of abstraction ($\mathbb{A}_{l\leftrightarrow h}$). To verify this, we compute the class membership among \EM{EM}, \TU{TU} and \QD{QD} for $600$ sketches using the softmax normalised distribution as: $\mathbb{A}_{l}$ of class ($\hat{\mathbb{A}}_{l}$), $\mathbb{A}_{m}$ of class ($\hat{\mathbb{A}}_{m}$), and $\mathbb{A}_{h}$ of class ($\hat{\mathbb{A}}_{h}$). These $600$ sketches are taken from the unseen split of $20$ categories common across \EM{EM}, \TU{TU} and \QD{QD}, where each category has $10$ sketches. From \cref{fig: expected-abstraction}, while there is an expected peak near class labels $(\hat{\mathbb{A}}_{l}, \hat{\mathbb{A}}_{m}, \hat{\mathbb{A}}_{h})$, we observe: (i) a significant number of sketches lie in the continuous spectrum between $\hat{\mathbb{A}}_{l} \leftrightarrow \hat{\mathbb{A}}_{m}$ and $\hat{\mathbb{A}}_{m} \leftrightarrow \hat{\mathbb{A}}_{h}$. This indicates that sketch abstraction is \textit{not discrete} at $\hat{\mathbb{A}}_{l}, \hat{\mathbb{A}}_{m}, \hat{\mathbb{A}}_{h}$ but \textit{continuous} from \EM{EM} $\leftrightarrow$ \QD{QD}. (ii) The abstraction of sketches in \TU{TU} vary widely, overlapping with those in \QD{QD} and \EM{EM}. Assigning all sketches in \TU{TU} (class $\hat{\mathbb{A}}_{m}$) to a fixed discrete level modelled by $\theta_{m}$ can corrupt generalisation \cite{zhang2018mixup}. To alleviate this hard assumption, we propose \textit{abstraction-mixup} -- a simple extension of an augmentation routine, \textit{mixup} \cite{zhang2018mixup, verma2019manifold}. Now, we randomly sample sketches $\{ \mathrm{I}^{l}_{S}, \mathrm{I}^{m}_{S}, \mathrm{I}^{h}_{S} \}$ from \{\QD{QD}, \TU{TU}, \EM{EM}\} and obtain $\{f^{l}_{s}, f^{m}_{s}, f^{h}_{s} \}$ (using $\mathrm{F}(\cdot)$) respectively. Next, we randomly sample the mixing coefficients from a $3$-dimensional Dirichlet distribution as $\{\lambda_{1}, \lambda_{2}, \lambda_{3} \} \sim \texttt{Dir}(\alpha)$ where,
\vspace{-2mm}
\begin{equation}
\begin{split}
    \texttt{Dir}(\lambda_{1}, & \lambda_{2},  \lambda_{3}; \alpha) = \frac{\Gamma(3 \alpha) }{ \Gamma(\alpha)^{3} } \prod_{i=1}^{3} \lambda_{i}^{\alpha - 1}
\end{split}
\end{equation}
\vspace{-4mm}

and $\Gamma(\cdot)$ is the gamma function with $\alpha > 0$. Next, we compute a \textit{mixup sketch feature} $f_{s}^{\alpha} = \lambda_{1}^{*} f_{s}^{l} + \lambda_{2}^{*} f_{s}^{m} + \lambda_{3}^{*} f_{s}^{h}$, where $\lambda_{i}^{*} = \lambda_{i} / (\sum_{j=1}^{3}\lambda_{j})$ is the normalised coefficients. Using $f^{\alpha}_{s}$, we train the codebook classifier $[\mathbb{A}_{l}^{\alpha}, \mathbb{A}_{m}^{\alpha}, \mathbb{A}_{h}^{\alpha}] = \mathcal{C}_{\theta}(f_{s}^{\alpha})$ by modifying \cref{eq: codebook-classify} as,
\vspace{-0.1cm}
\begin{equation}\label{eq: mixup}
    \mathcal{L}_\text{mix} = - (\lambda_{1}^{*} \log \mathbb{A}_{l}^{\alpha} + \lambda_{2}^{*} \log \mathbb{A}_{m}^{\alpha} + \lambda_{3}^{*} \log \mathbb{A}_{h}^{\alpha})
    \vspace{-2mm}
\end{equation}

\noindent Essentially, $\mathcal{L}_\text{mix}$ helps to augment synthetic latent representations of sketches across a continuous spectrum of \emph{sketch abstraction}. Our final training loss combines \cref{eq:clip-loss-meta-net,eq: raster-vector,eq: codebook-classify,eq: mixup} with coefficients (hyper-parameters) $\beta_{1,2,3}$ as,
\vspace{-0.1cm}
\begin{equation}
    \mathcal{L}_\text{tot} = \mathcal{L}_\text{CE} 
    + \beta_{1}\; \mathcal{L}_{\text{S} \rightarrow \text{V}} 
    + \beta_{2}\; \mathcal{L}_\text{CB} 
    + \beta_{3}\; \mathcal{L}_\text{mix}
    \vspace{-0.3cm}
\end{equation}

\begin{figure}[!t]
    \floatbox[{\capbeside\thisfloatsetup{capbesideposition={right,top},capbesidewidth=0.3\textwidth}}]{figure}[\FBwidth]
    {\includegraphics[trim={0.7cm 0 0 0},clip, width=0.7\textwidth]{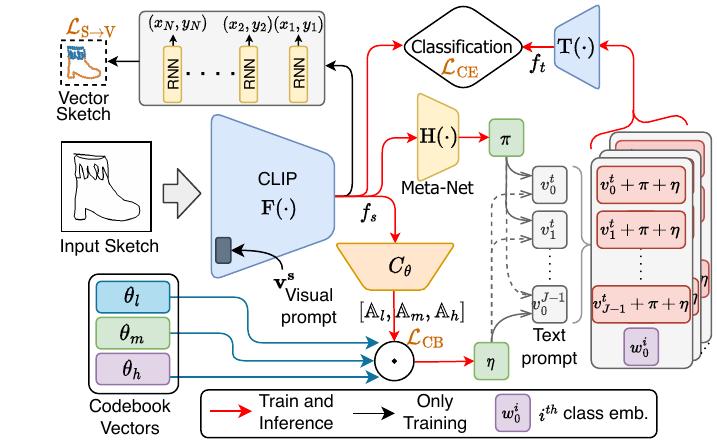}}
    {\caption[Caption beside image]{Given an input sketch, we compute visual feature $f_{s}$ with sketch prompts using a CLIP image encoder. Next, $f_{s}$ is fed into 4 pipelines: (i) An auxiliary raster-to-vector (sketch2vec) translation module that distils sketch-specific traits. (ii) A Meta-Net to predict an instance-specific context $\pi$ to generalise on unseen sketches. (iii) A codebook classifier $\mathcal{C}_{\theta}$ to
    
    }\label{fig:model_diagram}} \vspace{-0.03cm} 
    \justify
    compute an abstraction prompt $\eta$, given codebook vectors $\{\theta_{l}, \theta_{m}, \theta_{h}\}$. (iv) Combine text prompts $\mathbf{v^{t}}$ with $\pi$, $\eta$, and category embedding $w_{0}^{k}$ to generate text feature $f_{t}$ using CLIP text encoder $\mathrm{T}(\cdot)$. Finally, class-specific text features $f_{t}$ are matched with $f_{s}$ to compute class probabilities.
    \vspace{-0.7cm}
\end{figure}

\vspace{-0.4cm}
\subsection{Inference Pipleline}
\vspace{-0.1cm}
\textbf{First}, we use CLIP vision transformer $\mathrm{F}(\cdot)$ and our learned sketch prompts $\mathbf{v^{s}} = \{v^{s}_{0}, \dots, v^{s}_{J-1}\}$, where $v^{s}_{j} \in \mathbb{R}^{5 \times d_{p}}$, to encode an input sketch $\mathrm{I}_{S}$ into a visual feature $f_{s} = \mathrm{F}(\mathrm{I}_{S}; \mathbf{v^{s}}) \in \mathbb{R}^{d}$. \textbf{Second}, $f_{s}$ is simultaneously given to two modules: (i) a lightweight Meta-Net to predict instance-specific context, $\pi = \mathrm{H}(f_{s})$, where $\pi \in \mathbb{R}^{5 \times d_{t}}$, and (ii) a codebook classifier $\mathcal{C}_{\theta}: \mathbb{R}^{d} \to \mathbb{R}^{3}$ to get \texttt{softmax} normalised probability distribution over the three abstraction class labels, $[\mathbb{A}_{l}, \mathbb{A}_{m}, \mathbb{A}_{h}] = \mathcal{C}_{\theta}(f_{s})$. The predicted scores are used to get the abstraction prompt $\eta = \mathbb{A}_{l} \theta_{l} + \mathbb{A}_{m} \theta_{m} + \mathbb{A}_{h} \theta_{h}$, where $\eta \in \mathbb{R}^{5 \times d_{t}}$, and $\{ \theta_{l}, \theta_{m}, \theta_{h}\}$ are the codebook vectors. \textbf{Third}, for classification, we compute the word embedding for $K$ categories as $\{w_{0}^{k}\}_{k=1}^{K}$. \textbf{Finally}, we concatenate word embedding $w_{0}^{k}$ to the sum of our learned text prompt $\mathbf{v^{t}}$, instance-specific context $\pi$, and abstraction prompt $\eta$ to obtain the final text feature $f_{t}$ using CLIP text encoder as $f_{t} = \mathrm{T}( \ [ (\mathbf{v^{t}} + \pi + \eta), w_{0}^{k}] \ ) \in \mathbb{R}^{d}$. Classification probability is calculated from \cref{eq:clip-loss}.

\vspace{-0.4cm}
\section{Experiments}
\vspace{-0.3cm}

\keypoint{Implementation Details:} We use pre-trained CLIP with ViT-B/16 (vision transformer) as visual encoder $\mathrm{F}(\cdot)$ and transformer-based text encoder $\mathrm{T(\cdot)}$. For text encoder, we learn five $512$-dimensional context vectors as prompt $v^{t}_{j} \in \mathbb{R}^{5 \times 512}$. The class token for the $k$-th `\texttt{[category]}' is given by $w^{k}_{0} \in \mathbb{R}^{512}$. \cut{To adapt the vision encoder for sketch,} We learn five $768$-dimensional sketch prompts $v^{s}_{j} \in \mathbb{R}^{5 \times 768}$. The learnable prompts are injected upto a depth $J=9$, where $\{v^{s}_{0}, v^{t}_{0}\}$ are \textit{shallow} prompts, while the \textit{deep} prompts are $\{v^{s}_{1}, \dots, v^{s}_{J-1}\} \in \mathbb{R}^{8 \times 5 \times 512}$ and $\{v^{t}_{1}, \dots, v^{t}_{J-1}\} \in \mathbb{R}^{8 \times 5 \times 512}$ for vision and text encoders, respectively. Although CLIP's weights are frozen during training, we fine-tune the layer-norm parameters for improved performance \cite{sain2023clip}. Our method consisting of Meta-Net + Codebooks + layer-norm + vision prompts ($\mathrm{\mathbf{v^{s}}}$) + text prompts ($\mathrm{\mathbf{v^{t}}}$) is trained with Adam optimizer for $7$ epochs with  $1e-4$ learning rate and $64$ batch-size.

\vspace{0.1cm}
\keypoint{Datasets:} We use sketches from QuickDraw~\cite{quickdraw} and TU-Berlin~\cite{berlin} along with Edgemaps~\cite{chan2022learning} of TU-Berlin Extended~\cite{zhang2016sketchnet}. Ranked from highest to lowest abstraction, QuickDraw has $50$M sketches across $345$ categories, TU-Berlin has $20$K sketches from $250$ categories, and the Edgemaps are generated using \cite{chan2022learning} from {$204$K images} across $250$ categories in TU-Berlin extended. For few-shot training, we randomly pick $10$ sketches per class from a list of $125$ classes common to all three datasets and reserve the remaining ones ($220$ for QuickDraw, $125$ for TU-Berlin, and $125$ for Edgemaps) for zero-shot inference. Generating Edgemaps from complex scene images (with noisy backgrounds) leads to noisy sketches. We filter images with high classification scores (higher score $\Rightarrow$ less noisier background) using pre-trained zero-shot CLIP (details in supplementary).

\vspace{0.1cm}
\keypoint{Evaluation Setup:} We evaluate our algorithm on two fronts: \textit{(i) few-shot accuracy}: where we train our model on $10$ randomly sampled sketches from each of $125$ common classes in all three datasets and evaluate them on previously unseen samples from the same class list. \textit{(ii) zero-shot accuracy}: where we use our previously trained few-shot model and evaluate on unseen samples from new classes in these datasets. This difficult evaluation setup helps us understand \textit{(a)} how well the model generalises to unseen classes i.e., how much did we \textit{adapt} the generalisation potential of CLIP for sketch recognition, and \textit{(b)} how well the model trained on seen categories, generalises across varying abstractions using codebook vectors and their mix-up. We also evaluate the adaptability of our network, by replacing CLIP-backbone with FLAVA \cite{singh2022flava}.

\begin{table}[htbp]
    \RawFloats
    \centering
    \vspace{0.2cm}
    \begin{minipage}[t]{0.43\textwidth}
        \centering

            \tiny
             \scalebox{0.75}{
            \begin{tabular}{lcccccc}
            \toprule
            & \multicolumn{2}{c}{\EM{EM}} & \multicolumn{2}{c}{\TU{TU} \cite{berlin}} & \multicolumn{2}{c}{\QD{QD} \cite{quickdraw}}\\
            \cmidrule(lr){2-3}\cmidrule(lr){4-5}\cmidrule(lr){6-7}
            \multirow{-2}{*}{Methods} & Seen & Unseen & Seen & Unseen & Seen & Unseen\\ \midrule
            CLIP-Z \cite{clip} & 52.09 & 50.10 & 56.57 & 47.71 & 20.00 & 13.27 \\
            CoOp \cite{zhou2022learning} &55.06 & 50.66 &58.92 & 47.92&22.80 & 12.64 \\
            CoCoOp \cite{zhou2022conditional} &56.03 & 51.57&59.74 & 50.25&24.48 & 12.68 \\
            VPT-A \cite{bahng2022visual} & 52.82 & 41.22 & 66.08 & 51.01 & 37.02 & 15.36 \\
            MaPLe \cite{khattak2022maple} & 61.01 & 52.90 & 71.66 & 53.91 & 36.24 & 17.74\\
            Linear Probe \cite{clip} & 34.84 & -- & 57.24 & -- & 36.94 & -- \\
            Tip-Adapter \cite{tip-adapter2022} &60.58 &-- & 65.74 & -- & 42.24 & -- \\
            \cmidrule(lr){1-3}\cmidrule(lr){4-5}\cmidrule(lr){6-7}
            Sketch-A-Net \cite{yu2015sketch} & -- & -- & 27.01 & 3.14 & 18.08 & 0.68 \\
            ResNet \cite{he2016deep} & 9.00 & 2.16 & 18.34 & 1.82 & 7.20 & 0.63 \\
            ResNet-Adapt & 8.68 & 1.33 & 25.18 & 2.33 & 9.84 & 0.36 \\
            Edge-Augment & 14.95 & 0.70 & 35.17 & 0.81 & 9.52 & 0.46 \\
            \cmidrule(lr){1-3}\cmidrule(lr){4-5}\cmidrule(lr){6-7}
            VPT-P (Shallow) \cite{jia2022visual} & 27.92 & -- & 46.56 & - -& 24.33 & -- \\
            VPT-P (Deep) \cite{jia2022visual} & 42.08 & -- & 55.83 & -- & 34.00 & --\\
            \midrule
            \textbf{Ours} & \textbf{66.72} & \textbf{59.05} & \textbf{76.96} & \textbf{60.51} & \textbf{45.20} & \textbf{22.41} \\  \bottomrule
            \end{tabular}}
            \caption{\label{tab:acc} Recognition accuracy for sketches in \EM{EM}, \TU{TU}, and \QD{QD}. All competitors are jointly trained on \QD{QD} + \TU{TU} + \EM{EM}.}
    \end{minipage}
    \hfill
    \begin{minipage}[t]{0.53\textwidth}
        \centering
            \vspace{-1.7cm}
            \setlength{\tabcolsep}{3pt}
            \tiny
            \scalebox{0.75}{
            \begin{tabular}{ccccccccc}
            \toprule
            & & & & & & &\multicolumn{2}{c}{Accuracy}\\
            \multirow{-2}{*}{\makecell{Prompt\\Depth}} & \multirow{-2}{*}{\makecell{Context\\Token}} & \multirow{-2}{*}{\makecell{Meta\\ Net}} & \multirow{-2}{*}{\makecell{Layer\\Norm}} & \multirow{-2}{*}{\makecell{Codebook\\Vectors}} & \multirow{-2}{*}{Mixup} & \multirow{-2}{*}{\makecell{Sketch2\\Vec}} & Seen & Unseen \\\midrule
            1 & 5& \cmark& \cmark& \cmark& \cmark& \cmark&66.87 &59.39 \\
            3 & 5& \cmark& \cmark& \cmark& \cmark& \cmark&69.42 &58.99 \\
            7 & 5& \cmark& \cmark& \cmark& \cmark& \cmark&74.52 &57.77 \\
            \midrule
            9 & 2 & \cmark& \cmark& \cmark& \cmark& \cmark&74.52 &60.61 \\
            9 & 10 & \cmark& \cmark& \cmark& \cmark& \cmark&74.21 &57.56 \\
            9 & 20 & \cmark& \cmark& \cmark& \cmark& \cmark&75.54 & 55.94 \\
            \midrule
            9 & 5& \red{\xmark}& \cmark& \cmark& \cmark& \cmark&73.41 &54.19 \\
            9 & 5& \cmark& \red{\xmark}& \cmark& \cmark& \cmark&73.50 &59.90 \\
            9 & 5& \cmark& \cmark& \red{\xmark}& \red{\xmark}& \cmark&73.09 &60.20 \\
            9 & 5& \cmark& \cmark& \cmark & \red{\xmark}& \cmark&71.36 &60.00 \\
            9 & 5& \cmark& \cmark& \cmark & \cmark& \red{\xmark}&73.70 &59.49 \\
            \midrule
            {9} & {5} & \cmark& \cmark& \cmark& \cmark& \cmark& \textbf{76.96} & \textbf{60.51}\\\bottomrule
            \end{tabular}}
            \vspace{0.1cm}
            \caption{\label{tab:abl} Ablation studies on TU-Berlin \cite{berlin}: With varying \textit{Prompt Depth} and \textit{number of Context Tokens} trained with/without 
        \textit{LayerNorm fine-tuning},\textit{Meta-Net},  \textit{Codebook Vectors}, \textit{auxiliary Sketch2Vec} and \textit{Mixup}.}
            \label{tab:ablation2}
    \end{minipage}
    \vspace{-0.7cm}
\end{table}

\vspace{0.1cm}
\keypoint{Competitors:} We compare against \textit{(i)} existing state-of-the-art (SOTA) zero-shot and few-shot recognition methods. \textbf{CLIP} \cite{clip} classifies sketches by comparing a sketch encoding from the visual encoder with class encodings from the text encoder using hand-crafted text prompts like `\texttt{a photo of a [category]}'. \textbf{CoOp} \cite{zhou2022learning} extends {CLIP} by replacing hand-crafted prompts with learnable text prompts. \textbf{VPT-A} \cite{bahng2022visual} learns a visual prompt instead, that is added to the image to adapt the vision encoder for sketch classification via hand-crafted text prompts (similar to \textit{CLIP}). \textbf{MaPLe} \cite{khattak2022maple} learns a joint vision-text ``deep prompt'' inserted in multiple layers of the vision and text encoders for better sketch and class encodings respectively. We use the independent vision-language prompting mode of \textit{MaPLe} for a fairer comparison. Contrary to prior works learning static prompts, \textbf{CoCoOp} \cite{zhou2022conditional} learns an adaptive text prompt -- a bias vector $\pi$ conditioned on the input sketch and added to the learned text prompt in \textit{CoOp}. \textbf{Linear Probe} \cite{clip} classifies sketch by adding a linear layer at the end of {CLIP}'s visual encoder. 
\textbf{Tip-Adapter} \cite{zhang2022tip} uses a CLIP-based non-parametric query-key cache model~\cite{khandelwal2020} as an adapter with similarity-based retrieval to determine the class of test sample from its feature encoding. \textit{(ii)} Apart from adapting CLIP-based SOTAs, we provide a comprehensive comparison with widely used sketch classifiers like \textbf{ResNet}~\cite{he2016deep} and \textbf{Sketch-A-Net}~\cite{yu2015sketch}. \textbf{ResNet-Adapt} bridges the domain gap between abstract sketches in \TU{TU}~\cite{berlin}, \QD{QD} \cite{quickdraw}, and \EM{EM} using a domain discriminator to align ResNet visual features from all three domains. \textbf{Edge-Augment}~\cite{efthymiadis2022edge} fine-tunes a ResNet, pre-trained on geometrically augmented Edgemaps, on real sketches. \textbf{VPT-P}~\cite{jia2022visual} learns visual prompts for the vision transformer~\cite{dosovitskiy2021image}, injected as image patches in \textbf{VPT-P (Shallow)}, or multiple deeper layers \cite{khattak2022maple} in \textbf{VPT-P (Deep)}.

\vspace{-0.4cm}
\subsection{Sketch Recognition}
\vspace{-0.1cm}
\noindent We report few-shot and zero-shot recognition results of our algorithm on \QD{QD}, \TU{TU}, and \EM{EM} in \cref{tab:acc} using average accuracy across all datasets for reference.

\vspace{0.1cm}
\keypoint{Few-Shot Recognition:} We obtain a Top-1 accuracy of $66.72$\%, $76.96$\%, and $45.20$\% on \EM{EM}, \TU{TU}, and \QD{QD} respectively with our algorithm, beating SOTA \textit{MaPLe} by an average margin of $6.66$\%. Works on shallow language prompts like \textit{CoOp} ($45.59$\%) and \textit{CoCoOp} ($46.75$\%) and shallow visual prompts like \textit{VPT-A} ($51.97$\%) beat \textit{zero-shot CLIP} by an average of $2.71$\%, $3.87$\%, and $9.09$\%, reinforcing the idea of prompt learning for better recognition. We note that visual prompting on CLIP (\textit{VPT-A}) yields better results than language prompting (\textit{CoCoOp}) in a shallow prompt setting, with an accuracy difference of $5.22$\%. As analysed in \cite{khattak2022maple} and evident from our experiments, we note that deeper prompts like those in \textit{MaPLe} ($56.33$\%) and \textit{VPT-P (Deep)} ($43.97$\%) outperform their shallow prompt counterparts like \textit{VPT-P (shallow)} ($32.93$\%). Our work also outperforms few-shot methods that use hand-crafted prompts like \textit{Tip-Adapter} \cite{tip-adapter2022} and \textit{Linear Probe} \cite{clip} by $6.91$\% and $20.09$\% respectively. The superiority of our abstraction handling algorithm is demonstrated by its performance-\textit{gain} of $48.53$\% over naive adaptation-based solutions like \textit{ResNet-Adapt} ($14.56$\%) for multi-dataset training. \textit{Sketch-A-Net}, having a sketch-specific network architecture, beats \textit{ResNet/ResNet-Adapt} pre-trained on images by $8.67$/$1.83$\% and $10.88$/$8.24$\% on \TU{TU} and \QD{QD}, respectively. As \textit{Sketch-A-Net} requires stroke order information, we do not report accuracy against \EM{EM} that lack this data. 

\vspace{0.1cm}
\keypoint{Zero-Shot Recognition:} We find zero-shot performance in CLIP-based methods to be significantly higher than baseline networks, as CLIP is pre-trained on $400$M image-text pairs, making it easy for CLIP to  recognise sketch categories unseen during fine-tuning. Zero-shot CLIP beats non-CLIP baselines like \textit{Sketch-A-Net} and \textit{ResNet-Adapt} by as much as $29.27$\% and $35.69$\%, respectively. Amongst CLIP-based methods, we find our method tto outperform \textit{MaPLe} \cite{khattak2022maple} marginally ($5.80$\%) and \textit{zero-shot CLIP} considerably ($10.30$\%) in recognition accuracy. CLIP-based methods that are adapted specifically for \underline{few}-shot training (\textit{Tip-Adapter} \cite{tip-adapter2022} and \textit{Linear Probe} \cite{clip}) are not reported under \underline{Zero}-Shot Recognition. Furthermore, replacing the CLIP-backbone with FLAVA \cite{singh2022flava}, we find our model improves upon zero-shot recognition with FLAVA ($38.37$/$35.21$ \%) by $+17.36$/$+9.49$ \% for categories seen/unseen by our model.

\vspace{-0.4cm}
\subsection{Ablation}
\vspace{-0.1cm}
\noindent We ablate various components and hyperparameters in \cref{tab:abl}. \textit{(i)} Varying prompt depth $J$ affects recognition accuracy, where $J=1$ (shallow prompts) drops it to $66.87\%$ and $59.39\%$ in seen and unseen classes, respectively. \textit{(ii)} Varying length of language prompt ($v^{t}_{j} \in \mathbb{R}^{5 \times 512}$) from $5$ to $2$, $10$ and $20$ drops accuracy to $74.52$\%, $74.21$\% and $75.54$\% respectively. \textit{(ii)} Removing the Meta-Net drops accuracy by $4.94$\%, particularly zero-shot accuracy by $6.32$\%. \textit{(iii)} Removing\cut{ raster sketch $\rightarrow$ vector sketch loss} $\mathcal{L}_{\text{S} \rightarrow \text{V}}$ drops accuracy by $3.26\%$, due to lack of sketch-specific traits. \textit{(iv)} Removing codebook learning that models abstraction drops accuracy by $3.87\%$. \textit{(v)} Removing codebook mixup that models continuous abstraction drops accuracy by $3.06\%$. \textit{(vi)} Removing layer-norm drops accuracy by $2.03\%$.

\vspace{-0.3cm}
\subsection{Recognising the Degree of Abstraction}
\vspace{-0.1cm}

\noindent The codebook classifier predicts abstraction score as a softmax normalised probability distribution on $3$ coarse abstraction levels: $\hat{\mathbb{A}}_{l}$ (\EM{EM}), $\hat{\mathbb{A}}_{m}$ (\TU{TU}), $\hat{\mathbb{A}}_{h}$ (\QD{QD})\footnote{Since there is no precise abstraction annotation for each sketch, we assign coarse abstraction levels to \EM{EM}, \TU{TU}, and \QD{QD}.}.

\setlength{\intextsep}{1.5pt}
\begin{wrapfigure}{r}{0.5\textwidth}
\begin{tabular}{c|c}
        \begin{subfigure}[t]{0.5\linewidth}
          \centering
          \includegraphics[trim={0 0 0 4mm },clip,width=0.9\linewidth]{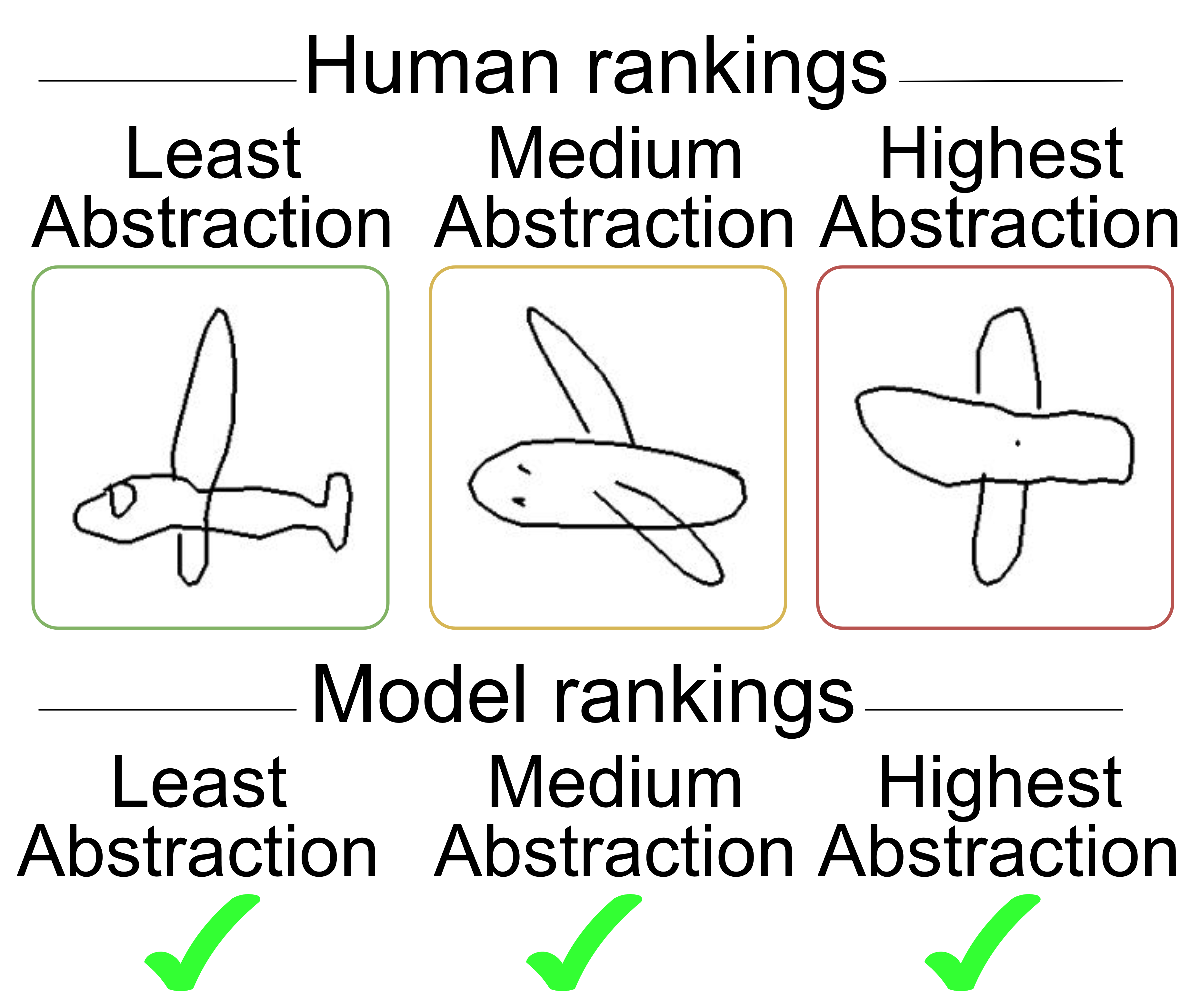}
          \caption{\tiny{Model agrees with human abstraction ranking}}
        \end{subfigure}
        &
        \begin{subfigure}[t]{0.5\linewidth}
          \centering
          \includegraphics[trim={4mm 2mm 0 0 },clip,width=0.9\linewidth]{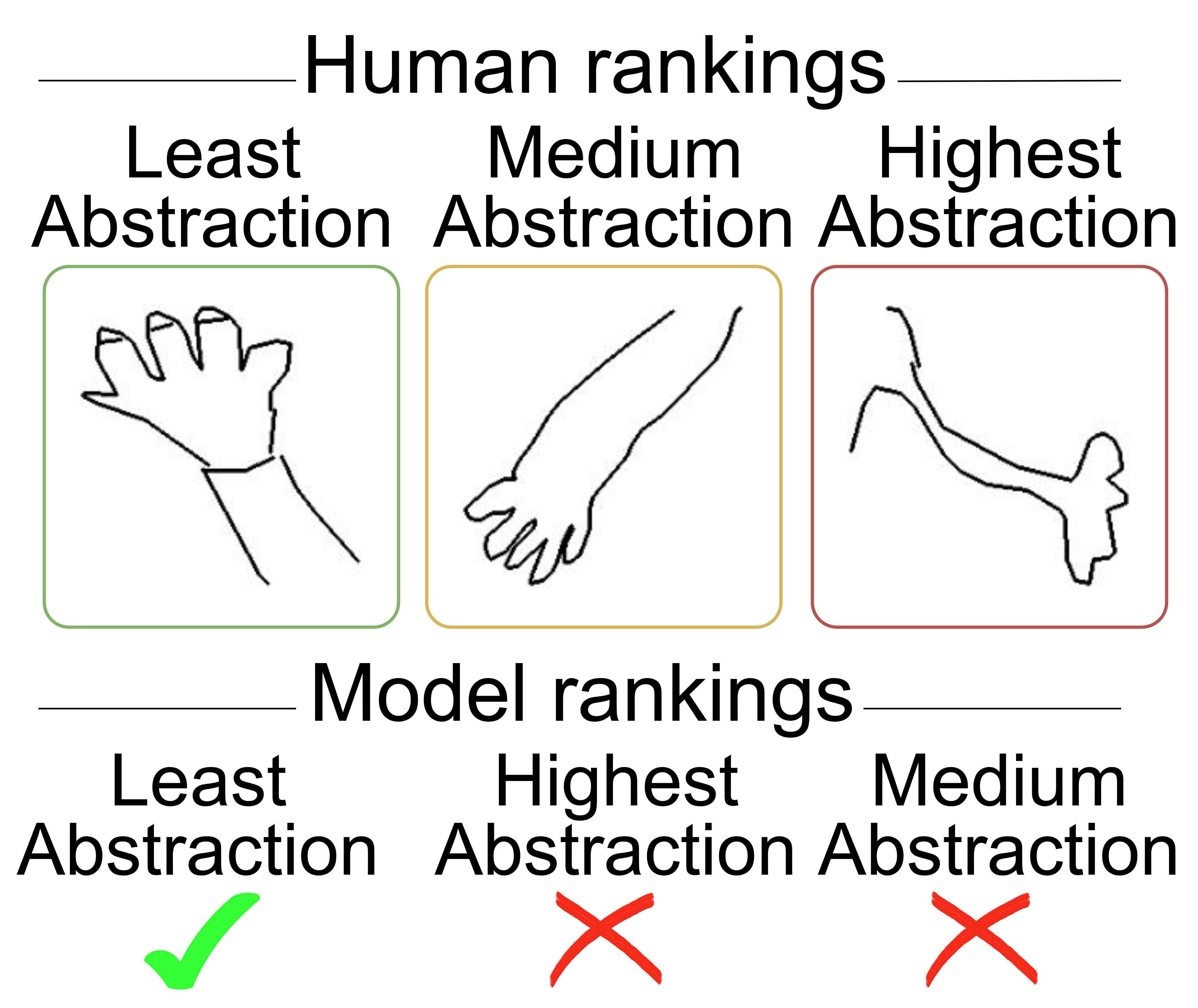}
          \caption{\tiny{Model disagrees with human abstraction ranking}}
        \end{subfigure}
\end{tabular}
\caption{Human study to rank $3$ sketches from same category and same dataset into low, medium, or high abstraction levels.}
\label{fig:human-study}
\vspace{-0.3cm}
\end{wrapfigure}

\keypoint{Quantitative:} Given unseen sketches from \QD{QD}, \TU{TU}, or \EM{EM}, we predict its abstraction score as a softmax normalised distribution over codebooks $\{\mathbb{A}_{l}, \mathbb{A}_{m}, \mathbb{A}_{h}\}$. Our model effectively classifies unseen sketches into their ground-truth abstraction level $\{\hat{\mathbb{A}}_{l}, \hat{\mathbb{A}}_{m}, \hat{\mathbb{A}}_{h}\}$ with an average accuracy of $95\%$. 
To verify that our abstraction classification is not merely a sketch dataset classification, we perform a human study, where we perform a detailed analysis of human rankings for abstraction levels $(0, 0.5, 1.0)$ against our predicted abstraction ranks.

\keypoint{Human Study:} Sketch abstraction is a subjective metric and hard to quantify. As prior works have no consensus on ``\textit{what is sketch abstraction}'' \cite{das2021sketchode, vinker2022clipasso, berlin, quickdraw}, we conduct a human study, where we select $20$ common categories across $3$ datasets.  Next from each dataset, and each category we select $3$ sketches (\textit{e.g.}, $3$ bicycle sketches all from TU-Berlin), giving us a total of $20\text{ (categories) } \times 3\text{ (datasets) } = 60$ triplets.

We then engage $5$ users across a diverse demographic in the age range of $20-30$\cut{, where each user has some level of experience with AI research}. We provide each user a set of $12$ triplets (a triplet has $3$ sketches per category per dataset) and ask them to rank each sketch (\cref{fig:human-study}) by its abstraction in the triplet.

Accordingly, the ranked-list of $60$ sketch triplets is compared with our model's prediction using the abstraction classifier. Our predicted ranking aligns with humans at an avg. correlation of $71.67$\%, $78.33$\%, and $66.67$\% for low, medium, and high abstraction levels respectively (\cref{tab:abs_sup}). 

\setlength{\intextsep}{1.5pt}

\begin{wrapfigure}{r}{0.6\textwidth}
\includegraphics[width=\linewidth]{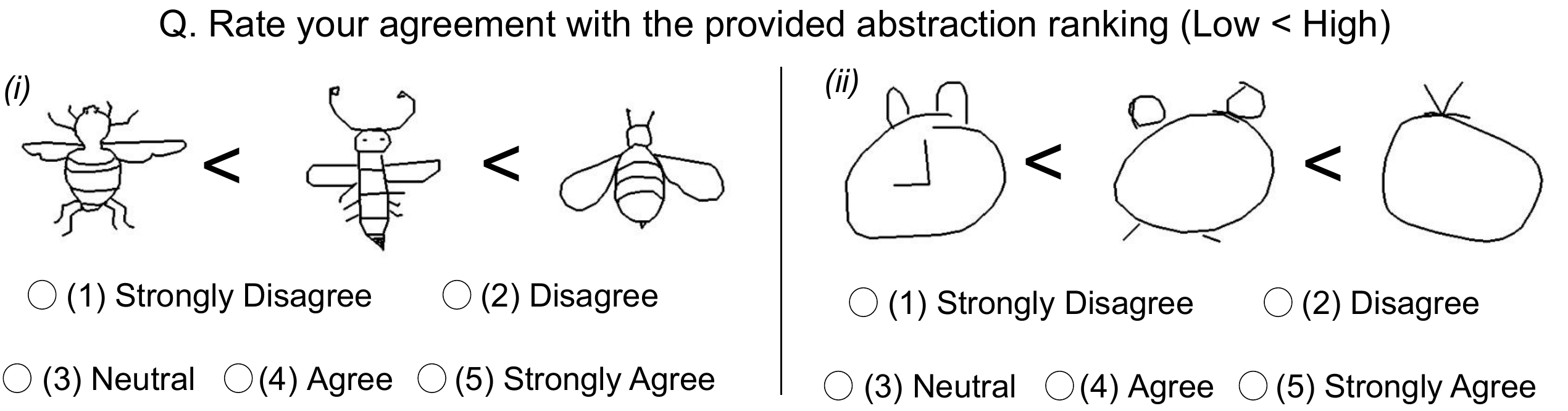}
\caption{User opinion on abstraction rankings}
\label{fig:rank_comp}
\vspace{-0.3cm}
\end{wrapfigure}

Next, we shuffle the $5$ sets of $12$ triplets among the $5$ users, such that no one receives their old set. Now, for each triplet, we show participants our predicted abstraction ranking and ask them to rate their overall agreement $(1-5)$ to the provided ranking (\cref{fig:rank_comp}) with $1$ being ``Strongly Disagree'' and $5$ being ``Strongly Agree''. For a total of $60$ triplets, users report an average agreement score (Mean Opinion Score) of $4.3$.
\begin{table}[htbp]
    \RawFloats
    \centering
    \vspace{0.2cm}
    \begin{minipage}[c]{0.5\textwidth}
        \includegraphics[width=\textwidth]{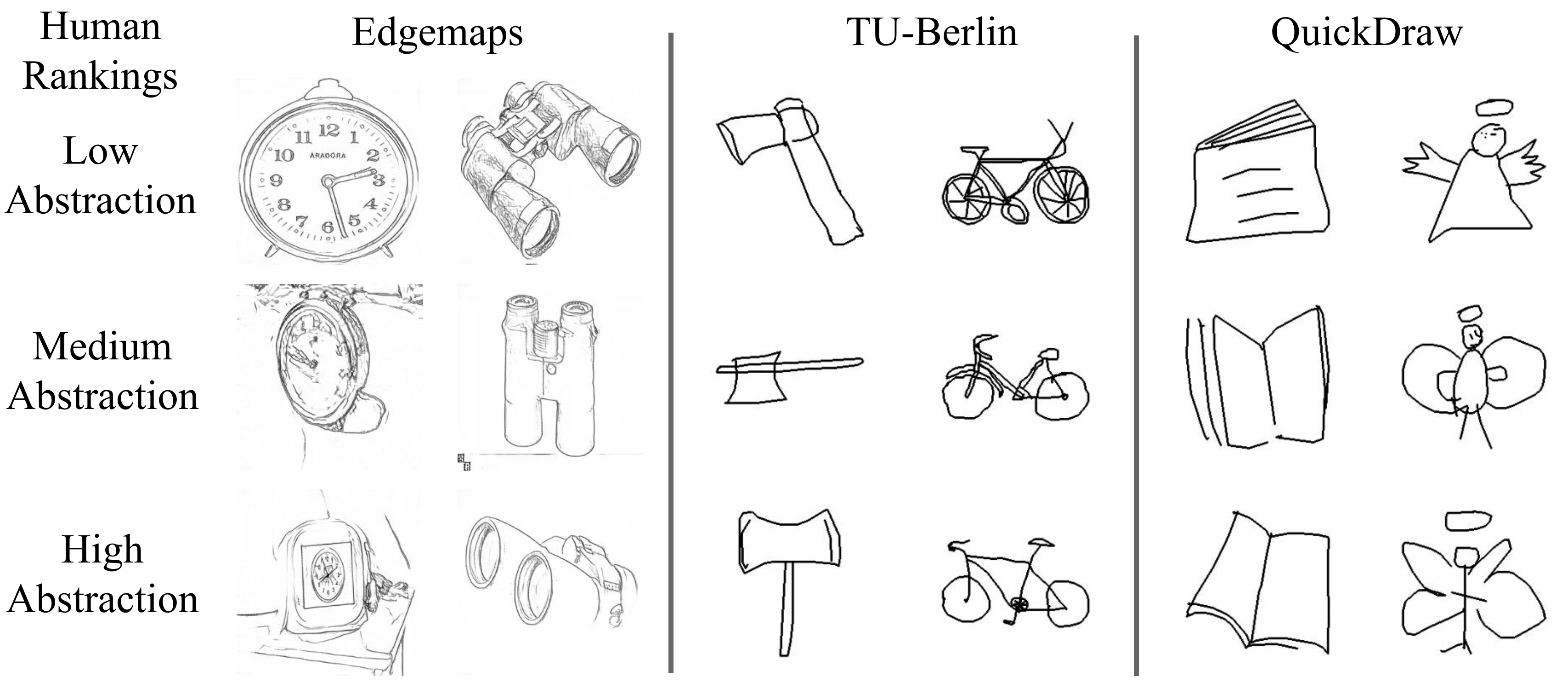}
        \captionof{figure}{Human rankings of sketch triplets.}
    \end{minipage}
    \hspace{0.3cm}
    \begin{minipage}[t]{0.4\textwidth}
        \centering
        \vspace{-1.7cm}
        \tiny
         \begin{tabular}{cccc}
         \toprule
         Human Study & \EM{EM} & \TU{TU} \cite{berlin} & \QD{QD} \cite{quickdraw}\\
         \midrule
         Rank-1 & 13 (65 \%) & 15 (75 \%)& 12 (60 \%)\\
         Rank-2 & 16 (80 \%)& 17 (85 \%)& 15 (75 \%)\\
         Rank-3 & 14 (70 \%)& 15 (75 \%)& 13 (65 \%)\\
         \bottomrule
         \label{tab:abs_sup}
         \end{tabular}
         \vspace{-0.25cm}
         \caption{No. of sketches on which our network agrees with the humans on most abstract (Rank-1), average (Rank-2), and least abstract (Rank-3).}       
    \end{minipage}
\end{table}
High correlation and Mean Opinion Scores, even when sketch triplets come from the same dataset, proves our codebook classifier is not simply a dataset classifier.
\vspace{-0.5cm}
\subsection{Evaluating On Unseen Abstraction Levels}
\vspace{-0.15cm}
\noindent 
To judge our model's generalisability, we simulate unseen sketch-abstraction levels using CLIPasso \cite{vinker2022clipasso}. Specifically, we test on unseen CLIPasso  generated sketches having $12, 16, 24$ and $32$ strokes representing decreasing order of abstraction. 

\begin{figure}
    \centering
    \includegraphics[width=0.93\linewidth]{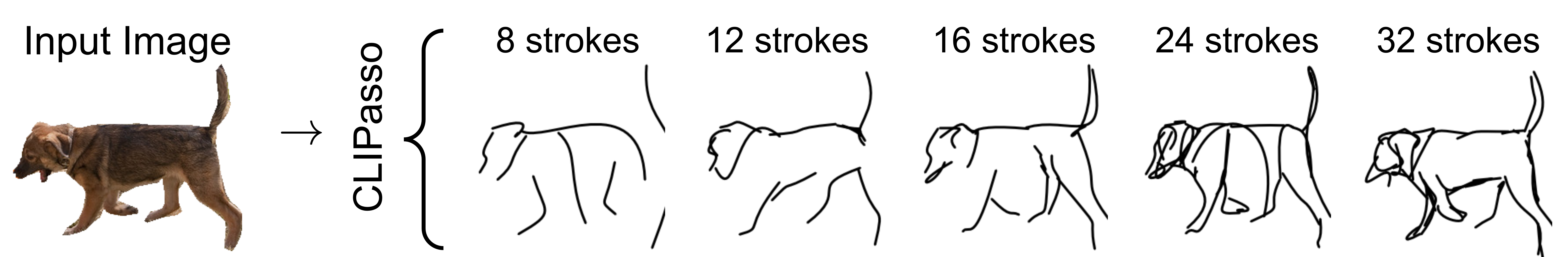}
    \vspace{0.3cm}
    \floatbox[{\capbeside\thisfloatsetup{capbesideposition={right,top},capbesidewidth=0.37\textwidth}}]{figure}[\FBwidth]
    {\vspace{0.1cm}
    \begin{subfigure}[t]{0.5\textwidth}
    {\includegraphics[width=\linewidth]{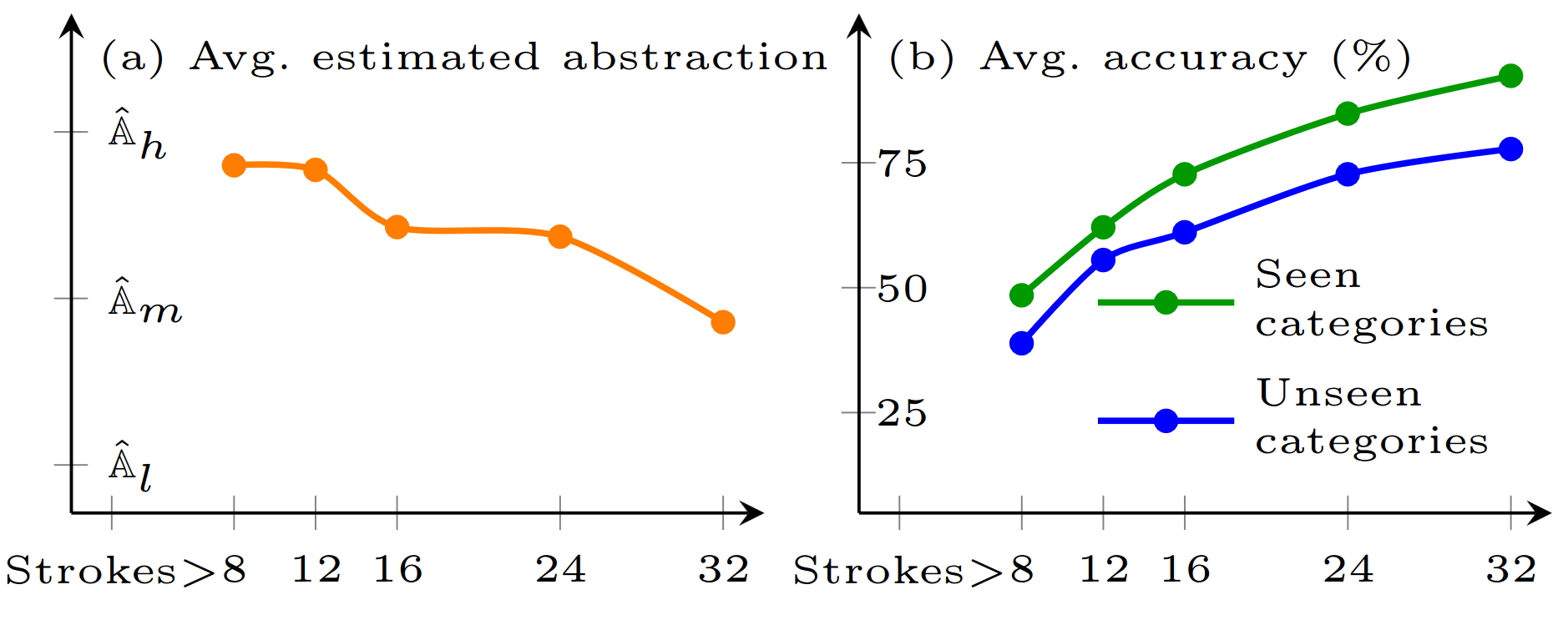}} 
    \end{subfigure}\vspace{0.1cm}
    }
    {\caption{\textbf{Top}: CLIPasso \cite{vinker2022clipasso} sketches at different number of strokes (and abstraction levels). \textbf{Left}: For CLIPasso, (a) Predicted abstraction scores vs. Num. strokes. (b) Sketch classification Acc. vs. Num. strokes}
    \label{fig:clipasso}}
    \vspace{-0.4cm}
    
\vspace{-4mm}
\end{figure}

\cref{fig:clipasso} shows our predicted abstraction scores (\textit{a}) to align with the notion of abstraction (\textit{c}) in CLIPasso (more strokes $\Rightarrow$ less abstract). 
A high recognition accuracy verifies our generalisability not only for abstraction prediction, but also for classification of \textit{unseen} sketches.

\vspace{-0.3cm}
\subsection{Interpreting Learned Abstraction Prompts}
\vspace{-0.15cm}
\noindent We aim to interpret the influence of abstraction prompt $\eta$ on learned text prompts ($\mathbf{v^{t}}$) to understand how it instils abstraction-aware knowledge into our sketch-classifier. Recent studies \cite{zhou2022learning} reveal that learned context tokens ($\mathbf{v^{t}}$) usually converge close to their initial embedding corresponding to the prompt of `\texttt{a sketch of a [category]}'. Influencing $\mathbf{v^{t}}$ with the codebook vector ($\eta$) however, pushes the embedding towards somewhat different words in the euclidean space \cite{khattak2022maple} which are sketch-specific in nature. For instance, when analysing sketches at \textit{lower} abstraction levels, we found cases where the euclidean distance from our prompt embedding to a word `artistic' is equivalent to that from other irrelevant words like `box', `camera', etc. This confusion dismisses naive measures like Euclidean distance as a tool for interpreting the influence of abstraction prompt $\eta$, necessitating further investigation for alternatives in future.

\vspace{-0.4cm}
\section{Conclusion}
\vspace{-0.3cm}
\noindent We extend the notion of a generalised classifier from photos to sketches. Towards this goal, we adapt CLIP (with open-set generalisation) for sketches by learning prompts for both the vision and language branches. In addition, to learn sketch-specific traits, we employ an auxiliary raster $\rightarrow$ vector sketch reconstruction loss. Finally, we generalise CLIP across varying sketch abstractions. As sketches lack precise abstraction annotation, we assign coarse-level scores to Edgemaps as low abstraction, TU-Berlin sketches  as medium, and QuickDraw doodles as high abstraction. We employ codebook learning and \textit{mixup} to learn sketch abstraction in a semi-supervised setup. The resulting SketchCLIP serves as a foundation model for robust sketch recognition algorithms based on large-scale vision-language models to classify \textit{``any''} abstract sketch.

\bibliographystyle{splncs04}
\bibliography{arxiv}

\begin{thebibliography}{10}
\providecommand{\url}[1]{\texttt{#1}}
\providecommand{\urlprefix}{URL }
\providecommand{\doi}[1]{https://doi.org/#1}

\bibitem{alaniz2022abstracting}
Alaniz, S., Mancini, M., Dutta, A., Marcos, D., Akata, Z.: {Abstracting Sketches through Simple Primitives}. In: ECCV (2022)

\bibitem{bahng2022visual}
Bahng, H., Jahanian, A., Sankaranarayanan, S., Isola, P.: {Exploring Visual Prompts for Adapting Large-Scale Models}. arXiv preprint arXiv:2203.17274  (2022)

\bibitem{baldrati2022effective}
Baldrati, A., Bertini, M., Uricchio, T., Del~Bimbo, A.: {Effective Conditioned and Composed Image Retrieval Combining CLIP-Based Features}. In: CVPR (2022)

\bibitem{open-world-recognition-2015}
Bendale, A., Boult, T.: {TowardsOpenWorldRecognition}. In: CVPR (2015)

\bibitem{berardi2023clip}
Berardi, G., Gryaditskaya, Y.: {Fine-Tuned but Zero-Shot 3D Shape Sketch View Similarity and Retrieval}. In: ICCV SHARP Workshop (ICCV) (2023)

\bibitem{berger2013style}
Berger, I., Shamir, A., Mahler, M., Carter, E., Hodgins, J.: Style and abstraction in portrait sketching. ACM TOG  (2013)

\bibitem{sketch2vec}
Bhunia, A.K., Chowdhury, P.N., Yang, Y., Hospedales, T.M., Xiang, T., Song, Y.Z.: {Vectorization and Rasterization: Self-Supervised Learning for Sketch and Handwriting}. In: CVPR (2021)

\bibitem{bhunia2020pixelor}
Bhunia, A.K., Das, A., Muhammad, U.R., Yang, Y., Hospedales, T.M., Xiang, T., Gryaditskaya, Y., Song, Y.Z.: {Pixelor: A Competitive Sketching AI Agent. So you think you can sketch?} ACM TOG  (2020)

\bibitem{bhunia2022doodle}
Bhunia, A.K., Gajjala, V.R., Koley, S., Kundu, R., Sain, A., Xiang, T., Song, Y.Z.: {Doodle it yourself: Class incremental learning by drawing a few sketches}. In: CVPR (2022)

\bibitem{brown2020language}
Brown, T., Mann, B., Ryder, N., Subbiah, M., Kaplan, J.D., Dhariwal, P., Neelakantan, A., Shyam, P., Sastry, G., Askell, A., Agarwal, S., Herbert-Voss, A., Krueger, G., Henighan, T., Child, R., Ramesh, A., Ziegler, D., Wu, J., Winter, C., Hesse, C., Chen, M., Sigler, E., Litwin, M., Gray, S., Chess, B., Clark, J., Berner, C., McCandlish, S., Radford, A., Sutskever, I., Amodei, D.: {Language Models are Few-Shot Learners}. In: NeurIPS (2020)

\bibitem{chan2022learning}
Chan, C., Durand, F., Isola, P.: {Learning to generate line drawings that convey geometry and semantics}. In: CVPR (2022)

\bibitem{chen2020deepfacedrawing}
Chen, S.Y., Su, W., Gao, L., Xia, S., Fu, H.: {DeepFaceDrawing: Deep generation of face images from sketches}. ACM TOG  (2020)

\bibitem{chen2018sketchygan}
Chen, W., Hays, J.: {Sketchygan: Towards diverse and realistic sketch to image synthesis}. In: ICCV (2018)

\bibitem{text2light2022}
Chen, Z., Wang, G., Liu, Z.: {Text2Light: Zero-Shot Text-Driven HDR Panorama Generation}. ACM TOG  (2022)

\bibitem{collomosse2019livesketch}
Collomosse, J., Bui, T., Jin, H.: {Livesketch: Query perturbations for guided sketch-based visual search}. In: CVPR (2019)

\bibitem{das2020beziersketch}
Das, A., Yang, Y., Hospedales, T., Xiang, T., Song, Y.Z.: {B{\'e}zierSketch: A generative model for scalable vector sketches}. In: ECCV (2020)

\bibitem{das2021sketchode}
Das, A., Yang, Y., Hospedales, T., Xiang, T., Song, Y.Z.: {SketchODE: Learning neural sketch representation in continuous time}. In: ICLR (2021)

\bibitem{deng2009imagenet}
Deng, J., Dong, W., Socher, R., Li, L.J., Li, K., Fei-Fei, L.: {Imagenet: A large-scale hierarchical image database}. In: CVPR (2009)

\bibitem{devlin2018bert}
Devlin, J., Chang, M.W., Lee, K., Toutanova, K.: {BERT: Pre-training of Deep Bidirectional Transformers for Language Understanding}. In: NAACL (2019)

\bibitem{dosovitskiy2021image}
Dosovitskiy, A., Beyer, L., Kolesnikov, A., Weissenborn, D., Zhai, X., Unterthiner, T., Dehghani, M., Minderer, M., Heigold, G., Gelly, S., Uszkoreit, J., Houlsby, N.: {An Image is Worth 16x16 Words: Transformers for Image Recognition at Scale}. In: ICLR (2021)

\bibitem{dutta2019semantically}
Dutta, A., Akata, Z.: {Semantically tied paired cycle consistency for zero-shot sketch-based image retrieval}. In: CVPR (2019)

\bibitem{efthymiadis2022edge}
Efthymiadis, N., Tolias, G., Chum, O.: {Edge Augmentation for Large-Scale Sketch Recognition without Sketches}. In: ICPR (2022)

\bibitem{berlin}
Eitz, M., Hays, J., Alexa, M.: {How Do Humans Sketch Objects?} ACM TOG  (2012)

\bibitem{fang2021clip2video}
Fang, H., Xiong, P., Xu, L., Chen, Y.: {Clip2video: Mastering video-text retrieval via image clip}. arXiv preprint arXiv:2106.11097  (2021)

\bibitem{gao2020sketchycoco}
Gao, C., Liu, Q., Xu, Q., Wang, L., Liu, J., Zou, C.: {Sketchycoco: Image generation from freehand scene sketches}. In: CVPR (2020)

\bibitem{CLIP-Adapter}
Gao, P., Geng, S., Zhang, R., Ma, T., Fang, R., Zhang, Y., Li, H., Qiao, Y.: {CLIP-Adapter: Better Vision-Language Models with Feature Adapters}. IJCV  (2023)

\bibitem{ge2021creative}
Ge, S., Goswami, V., Zitnick, L., Parikh, D.: {Creative Sketch Generation}. In: ICLR (2021)

\bibitem{gryaditskaya2019opensketch}
Gryaditskaya, Y., Sypesteyn, M., Hoftijzer, J.W., Pont, S., Durand, F., Bousseau, A.: {OpenSketch: A Richly-Annotated Dataset of Product Design Sketches}. ACM SIGGRAPH  (2019)

\bibitem{gu2022openvocabulary}
Gu, X., Lin, T.Y., Kuo, W., Cui, Y.: {Open-vocabulary Object Detection via Vision and Language Knowledge Distillation}. In: ICLR (2022)

\bibitem{guillard2021sketch2mesh}
Guillard, B., Remelli, E., Yvernay, P., Fua, P.: {Sketch2Mesh: Reconstructing and Editing 3D Shapes from Sketches}. In: CVPR (2021)

\bibitem{quickdraw}
Ha, D., Eck, D.: {A Neural Representation of Sketch Drawings}. In: ICLR (2018)

\bibitem{he2016deep}
He, K., Zhang, X., Ren, S., Sun, J.: {Deep residual learning for image recognition}. In: ICCV (2016)

\bibitem{hertzmann2020line}
Hertzmann, A.: {Why do line drawings work? A realism hypothesis}. Perception  (2020)

\bibitem{hu2020sketch}
Hu, C., Li, D., Yang, Y., Hospedales, T.M., Song, Y.Z.: {Sketch-a-segmenter: Sketch-based photo segmenter generation}. IEEE TIP  (2020)

\bibitem{jia2022visual}
Jia, M., Tang, L., Chen, B.C., Cardie, C., Belongie, S., Hariharan, B., Lim, S.N.: Visual prompt tuning. In: ECCV (2022)

\bibitem{scaling-laws}
Kaplan, J., McCandlish, S., Henighan, T., Brown, T.B., Chess, B., Child, R., Gray, S., Radford, A., Wu, J., Amodei, D.: {Scaling Laws for Neural Language Models}. arXiv preprint arXiv:2001.08361  (2020)

\bibitem{khandelwal2020}
Khandelwal, U., Levy, O., Jurafsky, Zettlemoyer, L., Lewis, M.: {Generalization through Memorization: Nearest Neighbor Language Models}. In: ICLR (2020)

\bibitem{khattak2022maple}
Khattak, M.U., Rasheed, H., Maaz, M., Khan, S., Khan, F.S.: {MaPLe: Multi-modal Prompt Learning}. In: CVPR (2023)

\bibitem{lei2021less}
Lei, J., Li, L., Zhou, L., Gan, Z., Berg, T.L., Bansal, M., Liu, J.: {Less is more: Clipbert for video-and-language learning via sparse sampling}. In: CVPR (2021)

\bibitem{hanhui_multistage}
Li, H., Jiang, X., Guan, B., Wang, R., Thalmann, N.M.: {Multistage Spatio-Temporal Networks for Robust Sketch Recognition}. IEEE TIP  (2022)

\bibitem{lin2020sketch}
Lin, H., Fu, Y., Xue, X., Jiang, Y.G.: {Sketch-bert: Learning sketch bidirectional encoder representation from transformers by self-supervised learning of sketch gestalt}. In: CVPR (2020)

\bibitem{liu2021deflocnet}
Liu, H., Wan, Z., Huang, W., Song, Y., Han, X., Liao, J., Jiang, B., Liu, W.: {Deflocnet: Deep image editing via flexible low-level controls}. In: CVPR (2021)

\bibitem{mirowski2022clip}
Mirowski, P., Banarse, D., Malinowski, M., Osindero, S., Fernando, C.: {Clip-clop: Clip-guided collage and photomontage}. arXiv preprint arXiv:2205.03146  (2022)

\bibitem{muhammad2019goal}
Muhammad, U.R., Yang, Y., Hospedales, T.M., Xiang, T., Song, Y.Z.: {Goal-driven sequential data abstraction}. In: CVPR (2019)

\bibitem{muhammad2018learning}
Muhammad, U.R., Yang, Y., Song, Y.Z., Xiang, T., Hospedales, T.M.: {Learning deep sketch abstraction}. In: ICCV (2018)

\bibitem{VQ-VAE2017}
Oord, A.v.d., Vinyals, O., Kavukcuoglu, K.: {NeuralDiscreteRepresentationLearning}. In: NeurIPS (2017)

\bibitem{sketch-jigsaw}
Pang, K., Yang, Y., Hospedales, T.M., Xiang, T., Song, Y.Z.: {Solving Mixed-modal Jigsaw Puzzle for Fine-Grained Sketch-Based Image Retrieval}. In: CVPR (2020)

\bibitem{paleosketch}
Paulson, B., Hammond, T.: {PaleoSketch: Accurate Primitive Sketch Recognition and Beautification}. In: IUI (2008)

\bibitem{petroni2019prompt}
Petroni, F., Rockt\:aschel, T., Lewis, P., Bakhtin, A., Wu, Y., Miller, A.H., Riedel, S.: {Language models as knowledge bases?} In: EMNLP (2019)

\bibitem{sketchlattice}
Qi, Y., Su, G., Chowdhury, P.N., Li, M., Song, Y.Z.: {SketchLattice: Latticed Representation for Sketch Manipulation}. In: ICCV (2021)

\bibitem{clip}
Radford, A., Kim, J.W., Hallacy, C., Ramesh, A., Goh, G., Agarwal, S., Sastry, G., Askell, A., Mishkin, P., Clark, J., et~al.: {Learning transferable visual models from natural language supervision}. In: ICML (2021)

\bibitem{DenseCLIP}
Rao, Y., Zhao, W., Chen, G., Tang, Y., Zhu, Z., Huang, G., Zhou, J., Lu, J.: {DenseCLIP: Language-Guided Dense Prediction with Context-Aware Prompting}. In: CVPR (2022)

\bibitem{ribeiro2020sketchformer}
Ribeiro, L.S.F., Bui, T., Collomosse, J., Ponti, M.: {Sketchformer: Transformer-based representation for sketched structure}. In: CVPR (2020)

\bibitem{LDM}
Rombach, R., Blattmann, A., Lorenz, D., Esser, P., Ommer, B.: {High-Resolution Image Synthesis with Latent Diffusion Models}. In: CVPR (2022)

\bibitem{sain2023clip}
Sain, A., Bhunia, A.K., Chowdhury, P.N., Koley, S., Xiang, T., Song, Y.Z.: {Clip for all things zero-shot sketch-based image retrieval, fine-grained or not}. In: CVPR (2023)

\bibitem{sain2020cross}
Sain, A., Bhunia, A.K., Yang, Y., Xiang, T., Song, Y.Z.: {Cross-Modal Hierarchical Modelling forFine-Grained Sketch Based Image Retrieval}. In: BMVC (2020)

\bibitem{sain2021stylemeup}
Sain, A., Bhunia, A.K., Yang, Y., Xiang, T., Song, Y.Z.: {Stylemeup: Towards style-agnostic sketch-based image retrieval}. In: CVPR (2021)

\bibitem{sangkloy2016sketchy}
Sangkloy, P., Burnell, N., Ham, C., Hays, J.: {The sketchy database: learning to retrieve badly drawn bunnies}. ACM TOG  (2016)

\bibitem{ravi2015recognition}
Sarvadevabhatla, R.K., Babu, R.V.: {Freehand Sketch Recognition Using Deep Features}. In: ICIP (2015)

\bibitem{schneider2014sketch}
Schneider, R.G., Tuytelaars, T.: {Sketch classification and classification-driven analysis using fisher vectors}. ACM TOG  (2014)

\bibitem{deepsketch2015}
Seddati, O., Dupont, S., Mahmoudi, S.: {DeepSketch: Deep Convolutional Neural Networks for Sketch Recognition and Similarity Search}. In: CBMI (2015)

\bibitem{deepsketch2016}
Seddati, O., Dupont, S., Mahmoudi, S.: {DeepSketch 2: Deep Convolutional Neural Networks for Partial Sketch Recognition}. In: CBMI (2016)

\bibitem{sezgin2001interface}
Sezgin, T.M., Stahovich, T., Davis, R.: {Sketch Based Interfaces: Early Processing for Sketch Understanding}. In: PUI (2001)

\bibitem{shen2018zero}
Shen, Y., Liu, L., Shen, F., Shao, L.: {Zero-shot sketch-image hashing}. In: ICCV (2018)

\bibitem{singh2022flava}
Singh, A., Hu, R., Goswami, V., Couairon, G., Galuba, W., Rohrbach, M., Kiela, D.: Flava: A foundational language and vision alignment model. In: CVPR (2022)

\bibitem{sketchhealer}
Su, G., Qi, Y., Pang, K., Yang, J., Song, Y.Z.: Sketchhealer: A graph-to-sequence network for recreating partial human sketches. In: BMVC (2020)

\bibitem{tripathi2020sketch}
Tripathi, A., Dani, R.R., Mishra, A., Chakraborty, A.: {Sketch-guided object localization in natural images}. In: ECCV (2020)

\bibitem{vaswani2017attention}
Vaswani, A., Shazeer, N., Parmar, N., Uszkoreit, J., Jones, L., Gomez, A.N., Kaiser, L.u., Polosukhin, I.: Attention is all you need. In: NeurIPS (2017)

\bibitem{verma2019manifold}
Verma, V., Lamb, A., Beckham, C., Najafi, A., Mitliagkas, I., Lopez-Paz, D., Bengio, Y.: {Manifold mixup: Better representations by interpolating hidden states}. In: ICML (2019)

\bibitem{vinker2022clipasso}
Vinker, Y., Pajouheshgar, E., Bo, J.Y., Bachmann, R.C., Bermano, A.H., Cohen-Or, D., Zamir, A., Shamir, A.: Clipasso: Semantically-aware object sketching. ACM TOG  (2022)

\bibitem{wang2021sketchembednet}
Wang, A., Ren, M., Zemel, R.: {Sketchembednet: Learning novel concepts by imitating drawings}. In: ICML (2021)

\bibitem{wang2021sketch}
Wang, S.Y., Bau, D., Zhu, J.Y.: {Sketch your own gan}. In: CVPR (2021)

\bibitem{wang2022clip}
Wang, Z., Liu, W., He, Q., Wu, X., Yi, Z.: {Clip-gen: Language-free training of a text-to-image generator with clip}. arXiv preprint arXiv:2203.00386  (2022)

\bibitem{xing2022class}
Xing, Y., Wu, Q., Cheng, D., Zhang, S., Liang, G., Zhang, Y.: Class-aware visual prompt tuning for vision-language pre-trained model. arXiv preprint arXiv:2208.08340  (2022)

\bibitem{xu2018sketchmate}
Xu, P., Huang, Y., Yuan, T., Pang, K., Song, Y.Z., Xiang, T., Hospedales, T.M., Ma, Z., Guo, J.: {Sketchmate: Deep hashing for million-scale human sketch retrieval}. In: ICCV (2018)

\bibitem{xu2022multigraph}
Xu, P., Joshi, C.K., Bresson, X.: {Multi-Graph Transformer for Free-Hand Sketch Recognition}. IEEE TNNLS  (2022)

\bibitem{xu2022domain}
Xu, R., Han, Z., Hui, L., Qian, J., Xie, J.: {Domain Disentangled Generative Adversarial Network for Zero-Shot Sketch-Based 3D Shape Retrieval}. In: AAAI (2022)

\bibitem{yan2020interactive}
Yan, G., Chen, Z., Yang, J., Wang, H.: {Interactive liquid splash modeling by user sketches}. ACM TOG  (2020)

\bibitem{yang2021sketchaa}
Yang, L., Pang, K., Zhang, H., Song, Y.Z.: {Sketchaa: Abstract representation for abstract sketches}. In: CVPR (2021)

\bibitem{lan2020S3Net}
Yang, L., Sain, A., Li, L., Qi, Y., Zhang, H., Song, Y.Z.: {S3NET: Graph Representational Network For Sketch Recognition}. In: ICME (2020)

\bibitem{yi2022animating}
Yi, R., Ye, Z., Fan, R., Shu, Y., Liu, Y.J., Lai, Y.K., Rosin, P.L.: {Animating portrait line drawings from a single face photo and a speech signal}. In: ACM SIGGRAPH (2022)

\bibitem{yu2019free}
Yu, J., Lin, Z., Yang, J., Shen, X., Lu, X., Huang, T.S.: {Free-form image inpainting with gated convolution}. In: CVPR (2019)

\bibitem{yu2015sketch}
Yu, Q., Yang, Y., Liu, F., Song, Y.Z., Xiang, T., Hospedales, T.M.: {Sketch-a-net: A deep neural network that beats humans}. IJCV  (2017)

\bibitem{zeng2022sketchedit}
Zeng, Y., Lin, Z., Patel, V.M.: {Sketchedit: Mask-free local image manipulation with partial sketches}. In: CVPR (2022)

\bibitem{zhang2018mixup}
Zhang, H., Cisse, M., Dauphin, Y.N., Lopez-Paz, D.: {mixup: Beyond Empirical Risk Minimization}. In: ICLR (2018)

\bibitem{zhang2016sketchnet}
Zhang, H., Liu, S., Zhang, C., Ren, W., Wang, R., Cao, X.: {Sketchnet: Sketch classification with web images}. In: ICCV (2016)

\bibitem{tip-adapter2022}
Zhang, R., Fang, R., Gao, P., Zhang, W., Li, K., Dai, J., Qiao, Y., Li, H.: {Tip-Adapter: Training-free CLIP-Adapter for Better Vision-Language Modeling}. In: ECCV (2022)

\bibitem{zhang2022tip}
Zhang, R., Zhang, W., Fang, R., Gao, P., Li, K., Dai, J., Qiao, Y., Li, H.: {Tip-Adapter: Training-free Adaption of CLIP for Few-shot Classification}. In: ECCV (2022)

\bibitem{zhang2021sketch2model}
Zhang, S.H., Guo, Y.C., Gu, Q.W.: {Sketch2Model: View-aware 3d modeling from single free-hand sketches}. In: CVPR (2021)

\bibitem{zhou2022conditional}
Zhou, K., Yang, J., Loy, C.C., Liu, Z.: {Conditional prompt learning for vision-language models}. In: CVPR (2022)

\bibitem{zhou2022learning}
Zhou, K., Yang, J., Loy, C.C., Liu, Z.: {Learning to prompt for vision-language models}. IJCV  (2022)

\bibitem{zhu2022prompt}
Zhu, B., Niu, Y., Han, Y., Wu, Y., Zhang, H.: {Prompt-aligned Gradient for Prompt Tuning}. In: ICCV (2023)

\end{thebibliography}
\end{document}